\documentclass[10pt, final,finalsubmission,twocolumn]{IEEEtran}

\usepackage{times}
\usepackage{epsfig}
\usepackage{graphicx}
\usepackage{amsmath}
\usepackage{amssymb}
\usepackage{multirow}
\usepackage{bbm}
\usepackage{rotating}
\usepackage{array}
\usepackage{threeparttable}
\usepackage{epstopdf}
\usepackage{fancyhdr}

\usepackage[breaklinks=true,bookmarks=false]{hyperref}

\begin{document}

%%%%%%%%% TITLE
\title{Deep Visual Domain Adaptation: A Survey}

\author{Mei Wang, Weihong Deng\\
School of Information and Communication Engineering,\\
Beijing University of Posts and Telecommunications, Beijing, China.\\
{\tt\small wm0245@126.com, whdeng@bupt.edu.cn}
}

\maketitle
\thispagestyle{fancy}        
\fancyhead{}                     
\lhead{Manuscript accepted by Neurocomputing 2018}
\cfoot{}
\rhead{\thepage} 

\renewcommand{\headrulewidth}{0.4pt}
\renewcommand{\footrulewidth}{0.4pt}

%\cfoot{}
\pagestyle{fancy}
\fancyhead{}
\lhead{Manuscript accepted by Neurocomputing 2018}
\cfoot{}
\rhead{\thepage}  
\renewcommand{\headrulewidth}{0.4pt}
\renewcommand{\footrulewidth}{0.4pt}

%%%%%%%%% ABSTRACT
\begin{abstract}

Deep domain adaptation has emerged as a new learning technique to address the lack of massive amounts of labeled data. Compared to conventional methods, which learn shared feature subspaces or reuse important source instances with shallow representations, deep domain adaptation methods leverage deep networks to learn more transferable representations by embedding domain adaptation in the pipeline of deep learning. There have been comprehensive surveys for shallow domain adaptation, but few timely reviews the emerging deep learning based methods. In this paper, we provide a comprehensive survey of deep domain adaptation methods for computer vision applications with four major contributions. First, we present a taxonomy of different deep domain adaptation scenarios according to the properties of data that define how two domains are diverged. Second, we summarize deep domain adaptation approaches into several categories based on training loss, and analyze and compare briefly the state-of-the-art methods under these categories. Third, we overview the computer vision applications that go beyond image classification, such as face recognition, semantic segmentation and object detection. Fourth, some potential deficiencies of current methods and several future directions are highlighted.

\end{abstract}

%%%%%%%%% BODY TEXT
\section{INTRODUCTION}

Over the past few years, machine learning has achieved great success and has benefited real-world applications. However, collecting and annotating datasets for every new task and domain are extremely expensive and time-consuming processes, sufficient training data may not always be available. Fortunately, the big data era makes a large amount of data available for other domains and tasks. For instance, although large-scale labeled video databases that are publicly available only contain a small number of samples, statistically, the YouTube face dataset (YTF) consists of 3.4K videos. The number of labeled still images is more than sufficient \cite{Sohn2017Unsupervised}. Hence, skillfully using the auxiliary data for the current task with scarce data will be helpful for real-world applications.

\begin{figure*}[htbp]
\centering
\includegraphics[width=14cm]{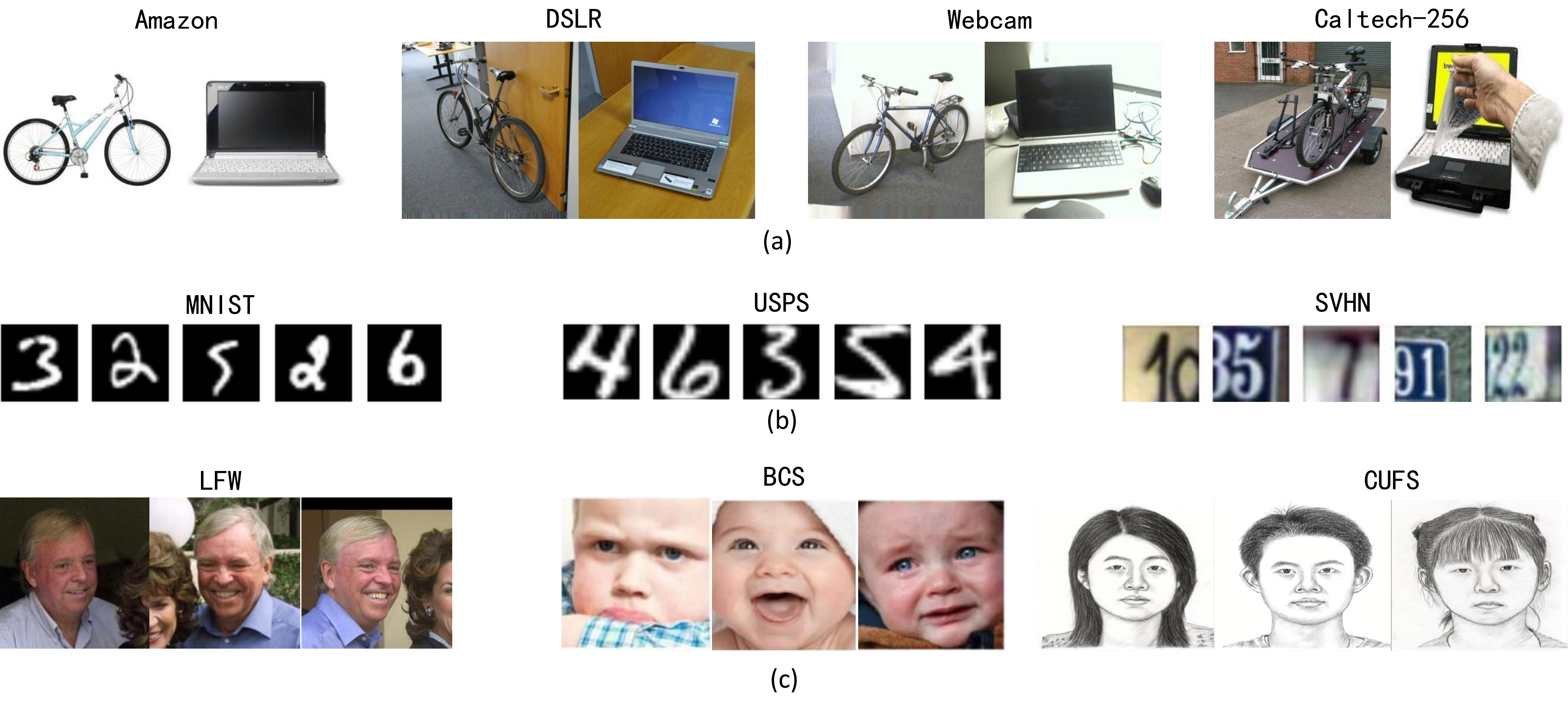}
\caption{(a) Some object images from the "Bike" and "Laptop" categories in Amazon, DSLR, Webcam, and Caltech-256 databases. (b) Some digit images from MNIST, USPS, and SVHN databases. (c) Some face images from LFW, BCS and CUFS databases. Realworld computer vision applications, such as face recognition, must learn to adapt to distributions specific to each domain.}
\label{fig10}
\end{figure*}

However, due to many factors (e.g., illumination, pose, and image quality), there is always a distribution change or domain shift between two domains that can degrade the performance, as shown in Fig. \ref{fig10}. Mimicking the human vision system, domain adaptation (DA) is a particular case of transfer learning (TL) that utilizes labeled data in one or more relevant source domains to execute new tasks in a target domain. Over the past decades, various shallow DA methods have been proposed to solve a domain shift between the source and target domains. The common algorithms for shallow DA can mainly be categorized into two classes: instance-based DA \cite{Bruzzone2010Domain,Chu2017Selective} and feature-based DA \cite{Gong2013Connecting,Pan2011Domain,gheisari2015unsupervised,pachori2017hashing}. The first class reduces the discrepancy by reweighting the source samples, and it trains on the weighted source samples. For the second class, a common shared space is generally learned in which the distributions of the two datasets are matched.

Recently, neural-network-based deep learning approaches have achieved many inspiring results in visual categorization applications, such as image classification \cite{Krizhevsky2012ImageNet}, face recognition \cite{Taigman2014DeepFace}, and object detection \cite{Girshick2013Rich}. Simulating the perception of the human brain, deep networks can represent high-level abstractions by multiple layers of non-linear transformations. Existing deep network architectures \cite{liu2017survey} include convolutional neural networks (CNNs) \cite{Krizhevsky2012ImageNet,Simonyan2014Very,Szegedy2015Going,He2016Deep}, deep belief networks (DBNs) \cite{Hinton2006A}, and stacked autoencoders (SAEs) \cite{Vincent2010Stacked}, among others. Although some studies have shown that deep networks can learn more transferable representations that disentangle the exploratory factors of variations underlying the data samples and group features hierarchically in accordance with their relatedness to invariant factors, Donahue et al. \cite{Donahue2013DeCAF} showed that a domain shift still affects their performance. The deep features would eventually transition from general to specific, and the transferability of the representation sharply decreases in higher layers. Therefore, recent work has addressed this problem by deep DA, which combines deep learning and DA.

There have been other surveys on TL and DA over the past few years \cite{Pan2010A,Shao2015Transfer,Day2017A,Patel2015Visual,Zhang2017Transfer,Csurka2017Domain}. Pan et al. \cite{Pan2010A} categorized TL under three subsettings, including inductive TL, transductive TL, and unsupervised TL, but they only studied homogeneous feature spaces. Shao et al. \cite{Shao2015Transfer} categorized TL techniques into feature-representation-level knowledge transfer and classifier-level knowledge transfer. The survey written by Patel \cite{Patel2015Visual} only focused on DA, a subtopic of TL. \cite{Day2017A} discussed 38 methods for heterogeneous TL that operate under various settings, requirements, and domains. Zhang et al. \cite{Zhang2017Transfer} were the first to summarize several transferring criteria in detail from the concept level. These five surveys mentioned above only cover the methodologies on shallow TL or DA.  The work presented by Csurka et al. \cite{Csurka2017Domain} briefly analyzed the state-of-the-art shallow DA methods and categorized the deep DA methods into three subsettings based on training loss: classification loss, discrepancy loss and adversarial loss. However, Csurka's work mainly focused on shallow methods, and it only discussed deep DA in image classification applications.

In this paper, we focus on analyzing and discussing deep DA methods. Specifically, the key contributions of this survey are as follows: 1) we present a taxonomy of different deep DA scenarios according to the properties of data that define how two domains are diverged. 2) extending Csurka's work, we improve and detail the three subsettings (training with classification loss, discrepancy loss and adversarial loss) and summarize different approaches used in different DA scenes. 3) Considering the distance of the source and target domains, multi-step DA methods are studied and categorized into hand-crafted, feature-based and representation-based mechanisms. 4) We provide a survey of many computer vision applications, such as image classification, face recognition, style translation, object detection, semantic segmentation and person re-identification.

The remainder of this survey is structured as follows. In Section II, we first define some notations, and then we categorize deep DA into different settings (given in Fig. \ref{fig1}). In the next three sections, different approaches are discussed for each setting, which are given in Table \ref{tab1} and Table \ref{tab2} in detail. Then, in Section VI, we introduce some successful computer vision applications of deep DA. Finally, the conclusion of this paper and discussion of future works are presented in Section VII.

\section{Overview}
\subsection{Notations and Definitions} \label{Notations and Definitions}

In this section, we introduce some notations and definitions that are used in this survey. The notations and definitions match those from the survey papers
by \cite{Pan2010A,Csurka2017Domain} to maintain consistency across surveys. A domain $\mathcal{D}$ consists of a feature space $\mathcal{X}$ and a marginal probability distribution $P(X)$, where $X=\{x_1,...,x_n\}\in\mathcal{X}$. Given a specific domain $\mathcal{D}=\{\mathcal{X},P(X)\}$, a task $\mathcal{T}$ consists of a feature space $\mathcal{Y}$ and an objective
predictive function $f(\cdot) $, which can also be viewed as a conditional probability distribution $P(Y|X)$ from a probabilistic perspective. In general, we can learn $P(Y|X)$ in a supervised manner from the labeled data $\{x_i,y_i\}$, where $x_i\in\mathcal{X}$ and $y_i\in\mathcal{Y}$.

Assume that we have two domains: the training dataset with sufficient labeled data is the source domain $\mathcal{D}^s=\{\mathcal{X}^s,P(X)^s\}$, and the test dataset with a small amount of labeled data or no labeled data is the target domain $\mathcal{D}^t=\{\mathcal{X}^t,P(X)^t\}$. We see that the partially labeled part, $\mathcal{D}^{tl}$, and the unlabeled parts, $\mathcal{D}^{tu}$, form the entire target domain, that is, $\mathcal{D}^t=\mathcal{D}^{tl}\cup\mathcal{D}^{tu}$. Each domain is together with its task: the former is $\mathcal{T}^s=\{\mathcal{Y}^s,P(Y^s|X^s)\}$,  and the latter is $\mathcal{T}^t=\{\mathcal{Y}^t,P(Y^t|X^t)\}$. Similarly, $P(Y^s|X^s)$ can be learned from the source labeled data $\{x_i^s,y_i^s\}$, while $P(Y^t|X^t)$ can be learned from labeled target data $\{x_i^{tl},y_i^{tl}\}$ and unlabeled data $\{x_i^{tu}\}$.

\begin{figure*}[htbp]
\centering
\includegraphics[width=14cm]{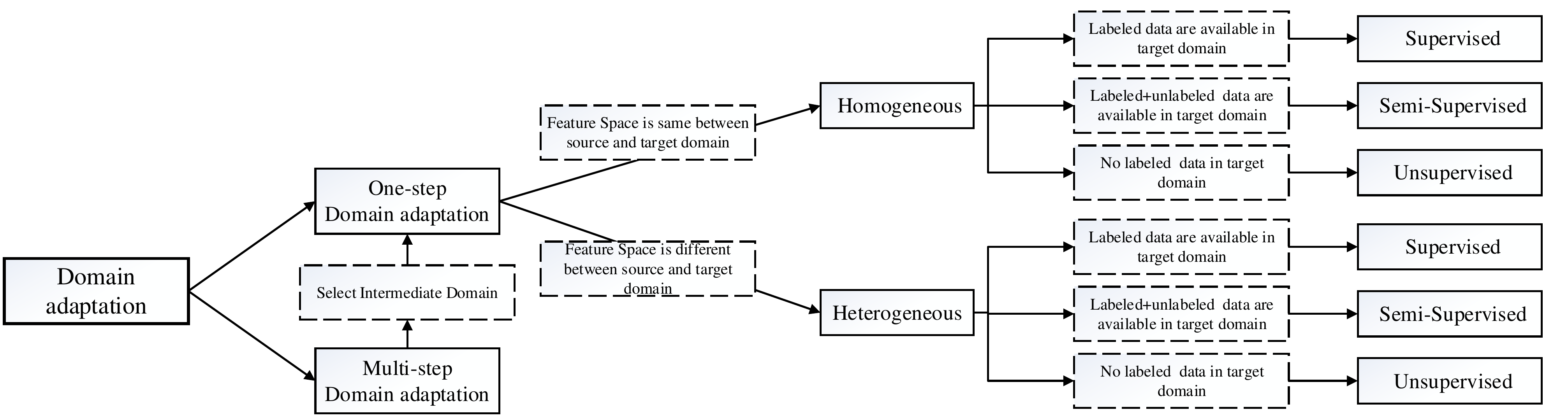}
\caption{An overview of different settings of domain adaptation}
\label{fig1}
\end{figure*}

\subsection{Different Settings of Domain Adaptation} \label{Settings of Domain Adaptation}

The case of traditional machine learning is $\mathcal{D}^s=\mathcal{D}^t$ and $\mathcal{T}^s=\mathcal{T}^t$. For TL, Pan et al. \cite{Pan2010A} summarized that the differences between different datasets can be caused by domain divergence $\mathcal{D}^s\neq\mathcal{D}^t$ (i.e., distribution shift or feature space difference) or task divergence $\mathcal{T}^s\neq\mathcal{T}^t$ (i.e., conditional distribution shift or label space difference), or both. Based on this summary, Pan et al. categorized TL into three main groups: inductive, transductive and unsupervised TL.

According to this classification, DA methods are transductive TL solutions with the assumption that the tasks are the same, i.e., $\mathcal{T}^s=\mathcal{T}^t$, and the differences are only caused by domain divergence, $\mathcal{D}^s\neq\mathcal{D}^t$. Therefore, DA can be split into two main categories based on different domain divergences (distribution shift or feature space difference): homogeneous and heterogeneous DA. Then, we can further categorize DA into supervised, semi-supervised and unsupervised DA in consideration of labeled data of the target domain. The classification is given in Fig. \ref{fig1}.

\begin{itemize}
\item In the \textbf{homogeneous DA} setting, the feature spaces between the source and target domains are identical ($\mathcal{X}^s=\mathcal{X}^t$) with the same dimension ($d^s=d^t$). Hence, the source and target datasets are generally different in terms of data distributions ($P(X)^s\neq P(X)^t$).
\end{itemize}

%\hangafter=0
%\setlength{\hangindent}{2em}
In addition, we can further categorize the homogeneous DA setting into three cases:

\begin{enumerate}
\item In the supervised DA, a small amount of labeled target data, $\mathcal{D}^{tl}$, are present. However, the labeled data are commonly not sufficient for tasks.
\item In the semi-supervised DA, both limited labeled data, $\mathcal{D}^{tl}$, and redundant unlabeled data, $\mathcal{D}^{tu}$, in the target domain are available in the training stage, which allows the networks to learn the structure information of the target domain.
\item In the unsupervised DA, no labeled but sufficient unlabeled target domain data, $\mathcal{D}^{tu}$, are observable when training the network.
\end{enumerate}
\begin{itemize}
\item In the \textbf{heterogeneous DA} setting, the feature spaces between the source and target domains are nonequivalent ($\mathcal{X}^s\neq\mathcal{X}^t$), and the dimensions may also generally differ ($d^s\neq d^t$).
\end{itemize}

%\hangafter=0
%\setlength{\hangindent}{2em}
Similar to the homogeneous setting, the heterogeneous DA setting can also be divided into supervised, semi-supervised and unsupervised DA.

\begin{figure*}[htbp]
\centering
\includegraphics[width=14cm]{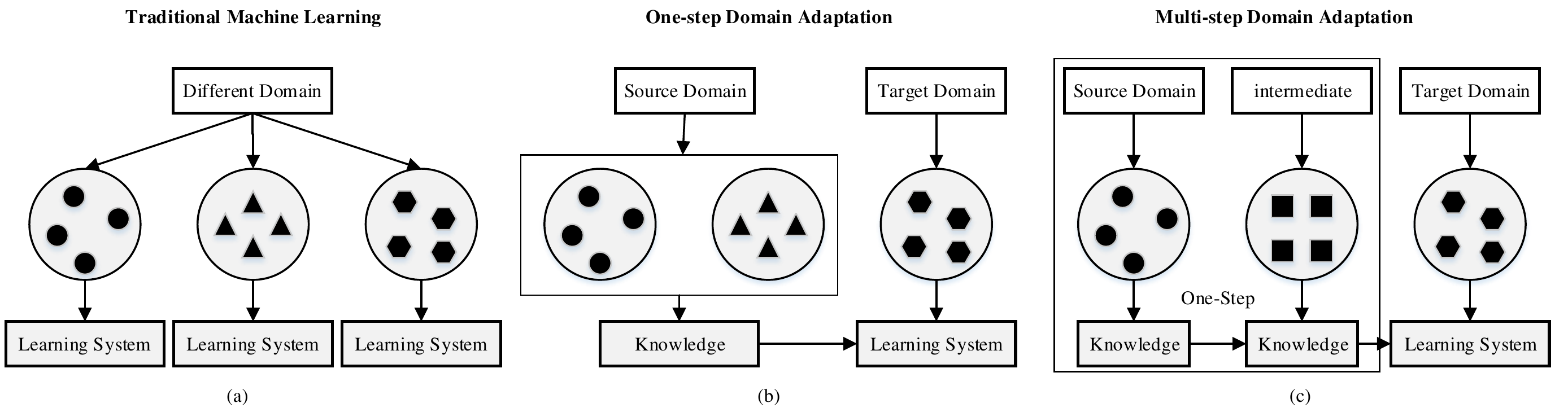}
\caption{Different learning processes between (a) traditional machine learning, (b) one-step domain adaptation and (c) multi-step domain adaptation \cite{Pan2010A}.}
\label{fig2}
\end{figure*}

All of the above DA settings assumed that the source and target domains are directly related; thus, transferring knowledge can be accomplished in one step. We
call them one-step DA. In reality, however, this assumption is occasionally unavailable. There is little overlap between the two domains, and performing one-step DA will not be effective. Fortunately, there are some intermediate domains that are able to draw the source and target domains closer than their original distance. Thus, we use a series of intermediate bridges to connect two seemingly unrelated domains and then perform one-step DA via this bridge, named multi-step (or transitive) DA \cite{Tan2015Transitive,tan2017distant}. For example, face images and vehicle images are dissimilar between each other due to different shapes or other aspects, and thus, one-step DA would fail. However, some intermediate images, such as 'football helmet', can be introduced to be an intermediate domain and have a smooth knowledge transfer. Fig. \ref{fig2} shows the differences between the learning processes of one-step and multi-step DA techniques.

\noindent

\section{Approaches of Deep Domain Adaptation}

\newcommand{\tabincell}[2]{\begin{tabular}{@{}#1@{}}#2\end{tabular}}  %表格自动换行

\begin{table*}[htbp]
	\centering

	\caption{Different Deep Approaches to One-Step DA}
	\begin{tabular}{|c|c|c|}
		\hline
        \tabincell{c}{One-step DA \\Approaches}          & Brief Description             & Subsettings \\ \hline
		 \multirow{4}[5]{2.2cm}{Discrepancy-based}& \multirow{4}[5]{6cm}{fine-tuning the deep network with labeled or unlabeled target data to diminish the domain shift} & \tabincell{c}{class criterion \cite{Tzeng2015Simultaneous,Peng2016Fine,motiian2017unified,saito2017asymmetric}\\
        \cite{Hu2015Deep,Hinton2015Distilling,Long2016Unsupervised,Zhang2015Deep,Yan2017Mind,Gebru2017Fine,Tzeng2015Simultaneous,Ge2017Borrowing} }        \\ \cline{3-3}
        &    &  \tabincell{c}{statistic criterion \cite{Long2016Deep,Yan2017Mind,Long2015Learning}\\ \cite{Long2016Unsupervised,Tzeng2014Deep,Ghifary2014Domain,Sun2016Deep,Peng2017Synthetic,Zhuang2015Supervised}}  \\ \cline{3-3}
         &    & \tabincell{c}{ architecture criterion \cite{Li2016Revisiting,Huang2017Arbitrary,Li2017Demystifying,Rozantsev2016Beyond,Xiao2016Learning,Rebuffi2017Learning}}\\ \cline{3-3}
        &   &  geometric criterion \cite{chopra2013dlid} \\ \hline
        \multirow{2}[2]{2.2cm}{Adversarial-based} &  \multirow{2}[2]{6cm}{using domain discriminators to encourage domain confusion through an adversarial objective}  & \tabincell{c}{ generative models  \cite{Liu2016Coupled,Bousmalis2016Unsupervised,Isola2016Image} } \\ \cline{3-3}
          &       & \tabincell{c}{ non-generative models \cite{Tzeng2017Adversarial,Tzeng2015Simultaneous,Ganin2017Domain,Ganin2015Unsupervised,Tzeng2015Adapting}\\ \cite{Peng2017Zero}}\\ \hline
        \multirow{2}[0]{2.2cm}{Reconstruction-based} &  \multirow{2}[0]{6cm}{using the data reconstruction as an auxiliary task to ensure feature invariance}    &   \tabincell{c}{encoder-decoder reconstruction  \cite{Bousmalis2016Domain,Ghifary2016Deep,Ghifary2015Domain,Zhuang2015Supervised} } \\ \cline{3-3}
        &      &   \tabincell{c}{adversarial reconstruction \cite{Yi2017DualGAN,Zhu2017Unpaired,Kim2017Learning}}    \\ \hline

	\end{tabular}
    \label{tab1}
\end{table*}

\begin{table*}[htbp]
\centering
\caption{Different Deep Approaches to Multi-Step DA}
 \begin{tabular}{|l|l|l|}
  \hline
   \tabincell{l}{Multi-step Approaches} & Brief Description \\ \hline
   Hand-crafted & \tabincell{l}{users determine the intermediate domains based on experience \cite{Xie2015Transfer}} \\ \hline
   Instance-based & \tabincell{l}{selecting certain parts of data from the auxiliary datasets to compose the intermediate \\domains \cite{tan2017distant,chopra2013dlid}} \\ \hline
   Representation-based & \tabincell{l}{freeze weights of one network and use their intermediate representations as input \\to the new network \cite{Rusu2016Progressive}} \\ \hline
 \end{tabular}
 \label{tab2}
\end{table*}

In a broad sense, deep DA is a method that utilizes a deep network to enhance the performance of DA. Under this definition, shallow methods with deep features \cite{Donahue2013DeCAF,Hoffman2013One,Raj2015Subspace,Nguyen2015DASH,zhang2017deep} can be considered as a deep DA approach. DA is adopted by shallow methods, whereas deep networks only extract vectorial features and are not helpful for transferring knowledge directly. For example, \cite{lu2017unsupervised} extracted the convolutional activations from a CNN as the tensor representation, and then performed tensor-aligned invariant subspace learning to realize DA. This approach reliably outperforms current state-of-the-art approaches based on traditional hand-crafted features because sufficient representational and transferable features can be extracted through deep networks, which can work better on discrimination tasks \cite{Donahue2013DeCAF}.

In a narrow sense, deep DA is based on deep learning architectures designed for DA and can obtain a firsthand effect from deep networks via back-propagation. The intuitive idea is to embed DA into the process of learning representation and to learn a deep feature representation that is both semantically meaningful and domain invariant. With the "good" feature representations, the performance of the target task would improve significantly. In this paper, we focus on the narrow definition and discuss how to utilize deep networks to learn "good" feature representations with extra training criteria.

\subsection{Categorization of One-Step Domain Adaptation}

In one-step DA, the deep approaches can be summarized into three cases, which refers to \cite{Csurka2017Domain}. Table 1 shows these three cases and brief descriptions. The first case is the discrepancy-based deep DA approach, which assumes that fine-tuning the deep network model with labeled or unlabeled target data can diminish the shift between the two domains. Class criterion, statistic criterion, architecture criterion and geometric criterion are four major techniques for performing fine-tuning.

\begin{itemize}
\item \textbf{Class Criterion:} uses the class label information as a guide for transferring knowledge between different domains. When the labeled samples from the target domain are available in supervised DA, soft label and metric learning are always effective \cite{Tzeng2015Simultaneous,Peng2016Fine,Hu2015Deep,Hinton2015Distilling,motiian2017unified}. When such samples are unavailable, some other techniques can be adopted to substitute for class labeled data, such as pseudo labels \cite{Long2016Unsupervised,Zhang2015Deep,Yan2017Mind,saito2017asymmetric} and attribute representation \cite{Gebru2017Fine,Tzeng2015Simultaneous}.
\end{itemize}

\begin{itemize}
\item \textbf{Statistic Criterion:} aligns the statistical distribution shift between the source and target domains using some mechanisms. The most commonly used methods for comparing and reducing distribution shift are maximum mean discrepancy (MMD) \cite{Long2016Deep,Yan2017Mind,Long2015Learning,Long2016Unsupervised,Tzeng2014Deep,Ghifary2014Domain}, correlation alignment (CORAL) \cite{Sun2016Deep,Peng2017Synthetic}, Kullback-Leibler (KL) divergence \cite{Zhuang2015Supervised} and $\mathcal{H}$ divergence, among others.
\end{itemize}

\begin{itemize}
\item \textbf{Architecture Criterion:} aims at improving the ability of learning more transferable features by adjusting the architectures of deep networks. The techniques that are proven to be cost effective include adaptive batch normalization (BN) \cite{Li2016Revisiting,Huang2017Arbitrary,Li2017Demystifying}, weak-related weight \cite{Rozantsev2016Beyond}, domain-guided dropout \cite{Xiao2016Learning}, and so forth.
\end{itemize}

\begin{itemize}
\item \textbf{Geometric Criterion:} bridges the source and target domains according to their geometrical properties. This criterion assumes that the relationship of geometric structures can reduce the domain shift \cite{chopra2013dlid}.
\end{itemize}

The second case can be referred to as an adversarial-based deep DA approach \cite{Ganin2017Domain}. In this case, a domain discriminator that classifies whether a data point is drawn from the source or target domain is used to encourage domain confusion through an adversarial objective to minimize the distance between the empirical source and target mapping distributions. Furthermore, the adversarial-based deep DA approach can be categorized into two cases based on whether there are generative models.

\begin{itemize}
\item \textbf{Generative Models:} combine the discriminative model with a generative component in general based on generative adversarial networks (GANs). One of the typical cases is to use source images, noise vectors or both to generate simulated samples that are similar to the target samples and preserve the annotation information of the source domain \cite{Liu2016Coupled,Bousmalis2016Unsupervised,Isola2016Image}.
\end{itemize}

\begin{itemize}
\item \textbf{Non-Generative Models:} rather than generating models with input image distributions, the feature extractor learns a discriminative representation using the labels in the source domain and maps the target data to the same space through a domain-confusion loss, thus resulting in the domain-invariant representations \cite{Tzeng2017Adversarial,Tzeng2015Simultaneous,Ganin2017Domain,Ganin2015Unsupervised,Tzeng2015Adapting}.
\end{itemize}

The third case can be referred to as a reconstruction-based DA approach, which assumes that the data reconstruction of the source or target samples can be helpful for improving the performance of DA. The reconstructor can ensure both specificity of intra-domain representations and indistinguishability of inter-domain representations.

\begin{itemize}
\item \textbf{Encoder-Decoder Reconstruction:} by using stacked autoencoders (SAEs), encoder-decoder reconstruction methods combine the encoder network for representation learning with a decoder network for data reconstruction \cite{Bousmalis2016Domain,Ghifary2016Deep,Ghifary2015Domain,Zhuang2015Supervised}.
\end{itemize}

\begin{itemize}
\item \textbf{Adversarial Reconstruction:} the reconstruction error is measured as the difference between the reconstructed and original images within each image domain by a cyclic mapping obtained via a GAN discriminator, such as dual GAN \cite{Yi2017DualGAN}, cycle GAN \cite{Zhu2017Unpaired} and disco GAN \cite{Kim2017Learning}.
\end{itemize}

\begin{table*}[htbp]
\small
\centering
\caption{Different Approaches used in Different Domain Adaptation Settings}
\begin{tabular}{|c|c|c|c|}
	\hline
     \multicolumn{2}{|c|}{}&  Supervised DA & Unsupervised DA \\ \hline
     \multirow{4}{*}{Discrepancy-based} & Class Criterion & $\surd$ & \\ \cline{2-4}
     & Statistic Criterion & & $\surd$ \\ \cline{2-4}
     & Architecture Criterion & $\surd$ & $\surd$ \\ \cline{2-4}
     &Geometric Criterion & $\surd$ &  \\ \hline
     \multirow{2}{*}{Adversarial-based} & Generative Model & & $\surd$\\ \cline{2-4}
     & Non-Generative Model & & $\surd$ \\ \hline
     \multirow{2}{*}{Reconstruction-based} & Encoder-Decoder Model & & $\surd$\\ \cline{2-4}
     & Adversarial Model& & $\surd$ \\ \hline
\end{tabular}
\label{tab3}
\end{table*}

\subsection{Categorization of Multi-Step Domain Adaptation}

In multi-step DA, we first determine the intermediate domains that are more related with the source and target domains than their direct connection. Second, the knowledge transfer process will be performed between the source, intermediate and target domains by one-step DA with less information loss. Thus, the key of multi-step DA is how to select and utilize intermediate domains; additionally, it can fall into three categories referring to \cite{Pan2010A}: hand-crafted, feature-based and representation-based selection mechanisms.

\begin{itemize}
\item \textbf{Hand-Crafted:} users determine the intermediate domains based on experience \cite{Xie2015Transfer}.
\end{itemize}

\begin{itemize}
\item \textbf{Instance-Based:} selecting certain parts of data from the auxiliary datasets to compose the intermediate domains to train the deep network \cite{tan2017distant,chopra2013dlid}.
\end{itemize}

\begin{itemize}
\item \textbf{Representation-Based:} transfer is enabled via freezing the previously trained network and using their intermediate representations as input to the new one \cite{Rusu2016Progressive}.
\end{itemize}
	
\section{ONE-STEP DOMAIN ADAPTATION}

As mentioned in Section \ref{Notations and Definitions}, the data in the target domain have three types regardless of homogeneous or heterogeneous DA: 1) supervised DA with labeled data, 2) semi-supervised DA with labeled and unlabeled data and 3) non-supervised DA with unlabeled data. The second setting is able to be accomplished by combining the methods of setting 1 and setting 3; thus, we only focus on the first and third settings in this paper. The cases where the different approaches are mainly used for each DA setting are shown in Table \ref{tab3}. As shown, more work is focused on unsupervised scenes because supervised DA has its limitations. When only few labeled data in the target domain are available, using the source and target labeled data to train parameters of models typically results in overfitting to the source distribution. In addition, the discrepancy-based approaches have been studied for years and produced more methods in many research works, whereas the adversarial-based and reconstruction-based approaches are a relatively new research topic but have recently been attracting more attention.

\subsection{Homogeneous Domain Adaptation}

\subsubsection{Discrepancy-Based Approaches} \label{Discrepancy-Supervised}

Yosinski et al.\cite{Yosinski2014How} proved that transferable features learned by deep networks have limitations due to fragile co-adaptation and representation specificity and that fine-tuning can enhance generalization performance. Fine-tuning (can also be viewed as a discrepancy-based deep DA approach) is to train a base network with source data and then directly reuse the first n layers to conduct a target network. The remaining layers of the target network are randomly initialized and trained with loss based on discrepancy. During training, the first n layers of the target network can be fine-tuned or frozen depending on the size of the target dataset and its similarity to the source dataset \cite{Chu2016Best}. Some common rules of thumb for navigating the 4 major scenarios are given in Table \ref{tab4}.

\begin{table*}[htbp]
\small
\centering
\caption{Some Common Rules of Thumb for Deciding Fine-tuned or Frozen in the First n Layers. \cite{Chu2016Best}}
\begin{tabular}{|c|c|c|c|c|}
	\hline
     & & \multicolumn{3}{|c|}{The Size of Target Dataset} \\ \cline{3-5}
     & & \textbf{Low} & \textbf{Medium} & \textbf{High} \\ \hline
     The Distance & \textbf{Low} & Freeze & Try Freeze or Tune & Tune \\ \cline{2-5}
     between & \textbf{Medium} & Try Freeze or Tune & Tune & Tune \\ \cline{2-5}
     Source and Target & \textbf{High} & Try Freeze or Tune & Tune & Tune \\ \hline
\end{tabular}
\label{tab4}
\end{table*}

\begin{figure}[htbp]
\centering
\label{figure：fig_ref_name}
\includegraphics[width=8cm]{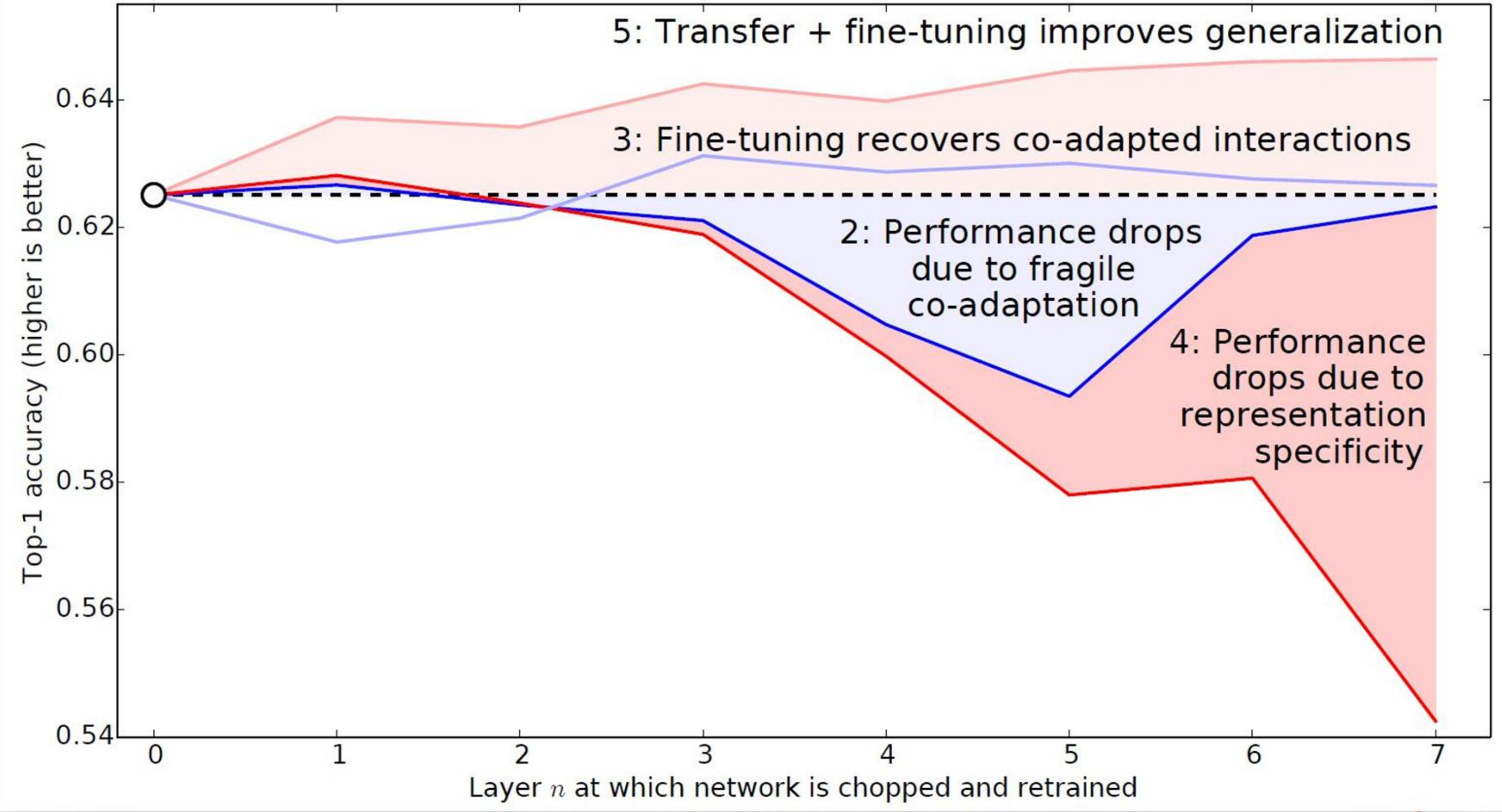}
\caption{The average accuracy over the validation set for a network trained with different strategies. Baseline B: the network is trained on dataset B. 2) BnB: the first n layers are reused from baseline B and frozen. The higher layers are trained on dataset B. 3) BnB+: the same as BnB but where all layers are fine-tuned. 4) AnB: the first n layers are reused from the network trained on dataset A and frozen. The higher layers are trained on dataset B. 5) AnB+: the same as AnB but where all layers are fine-tuned \cite{Yosinski2014How}.}
\end{figure}

\begin{itemize}
\item \textbf{Class Criterion}
\end{itemize}

The class criterion is the most basic training loss in deep DA. After pre-training the network with source data, the remaining layers of the target model use the class label information as a guide to train the network. Hence, a small number of labeled samples from the target dataset is assumed to be available.

Ideally, the class label information is given directly in supervised DA. Most work commonly uses the negative log-likelihood of the ground truth class with softmax as their training loss, $\mathcal{L}=-\sum\nolimits_{i=0}^{N}{{{y}_{i}}\log {{{\hat{y}}}_{i}}}$ ($ {{{\hat{y}}}_{i}}$ are the softmax
predictions of the model, which represent class probabilities) \cite{Tzeng2015Simultaneous,Peng2016Fine,Hu2015Deep,wang2016deep}. To extend this, Hinton et al. \cite{Hinton2015Distilling} modified the softmax function to soft label loss:
 \begin{equation}
{{q}_{i}}=\frac{\exp ({{z}_{i}}/T)}{\sum\nolimits_{j}{(\exp ({{z}_{j}}/T))}}
\end{equation}
where $z^i$ is the logit output computed for each class. $T$ is a temperature that is normally set to 1 in standard softmax, but it takes a higher value to produce a softer probability distribution over classes. By using it, much of the information about the learned function that resides in the ratios of very small probabilities can be obtained. For example, when recognizing digits, one version of 2 may obtain a probability of $10^6$ of being a 3 and $10^9$ of being a 7; in other words, this version of 2 looks more similar to 3 than 7. Inspired by Hinton, \cite{Tzeng2015Simultaneous} fine-tuned the network by simultaneously minimizing the domain confusion loss (belonging to adversarial-based approaches, which will be presented in Section \ref{Unsupervised-Adversarial}) and soft label loss. Using soft labels rather than hard labels can preserve the relationships between classes across domains. Gebru et al. \cite{Gebru2017Fine} modified existing adaptation algorithms based on \cite{Tzeng2015Simultaneous} and utilized soft label loss at the fine-grained class level $\mathcal{L}_{csoft}$ and attribute level $\mathcal{L}_{asoft}$.

\begin{figure}[htbp]
\centering
\label{figure：fig_ref_name}
\includegraphics[width=8cm]{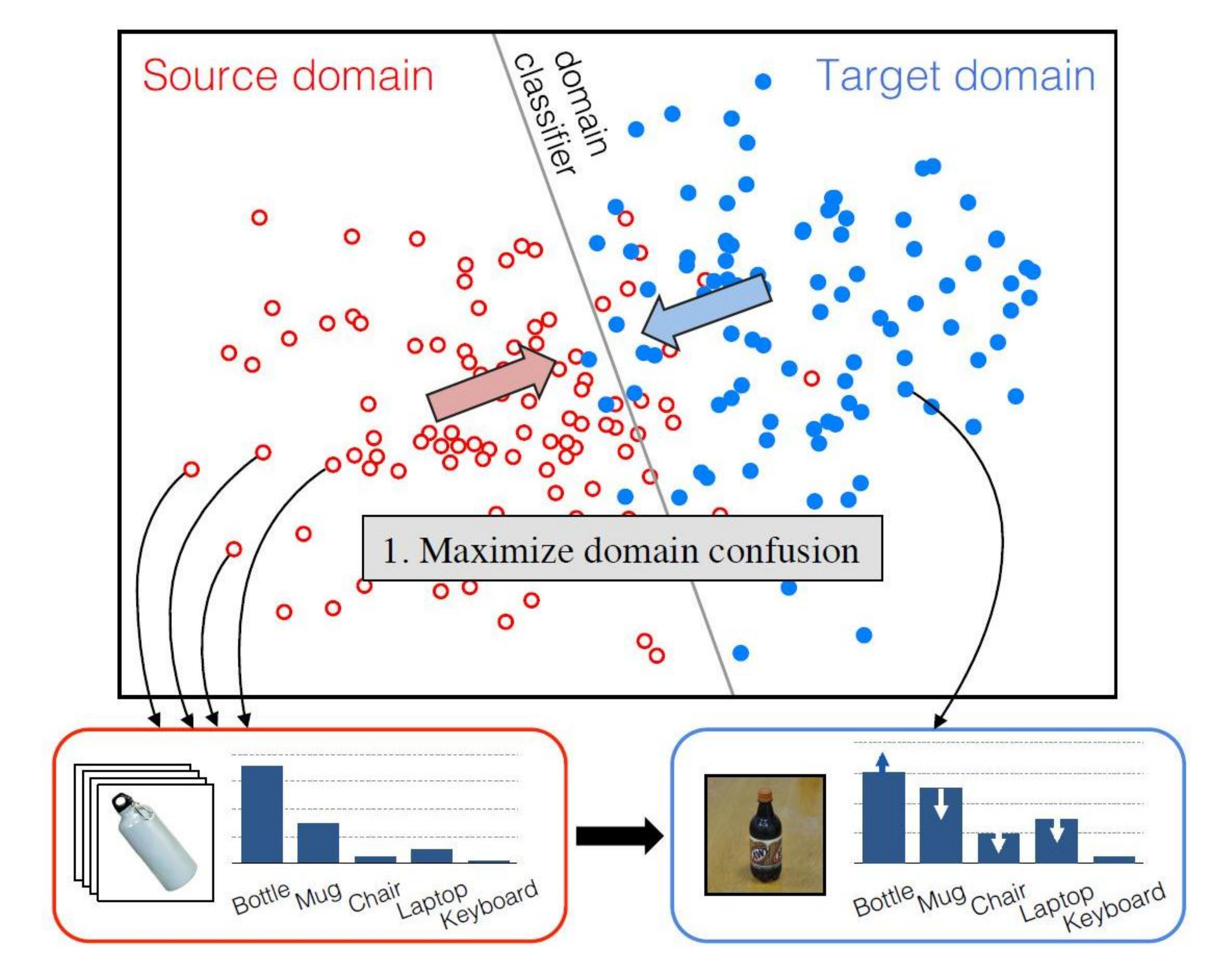}
\caption{Deep DA by combining domain confusion loss and soft label loss \cite{Tzeng2015Simultaneous}.}
\end{figure}

In addition to softmax loss, there are other methods that can be used as training loss to fine-tune the target model in supervised DA. Embedding metric learning in deep networks is another method that can make the distance of samples from different domains with the same labels be closer while those with different labels are far away. Based on this idea, \cite{motiian2017unified} constructed the semantic alignment loss and the separation loss accordingly. Deep transfer metric learning is proposed by \cite{Hu2015Deep}, which applies the marginal Fisher analysis criterion and MMD criterion (described in Statistic Criterion) to minimize their distribution difference:
\begin{equation}
\begin{aligned}
\min \mathcal{J} =S_{c}^{(M)}-\alpha S_{b}^{(M)}+\beta D_{ts}^{(M)}\left( {{\mathcal{X}}^{s}},{{\mathcal{X}}^{t}} \right) \\ +\gamma \sum\limits_{m=1}^{M}{(\left\| {{W}^{(m)}} \right\|_{F}^{2}+\left\| {{b}^{(m)}} \right\|_{2}^{2})}
\end{aligned}
\end{equation}
where $\alpha,\beta$ and $\gamma$ are regularization parameters and $W^{(m)}$ and $b^{(m)}$ are the weights and biases of the $m^{th}$ layer of the network. $D_{ts}^{(M)}\left( {{\mathcal{X}}^{s}},{{\mathcal{X}}^{t}} \right)$ is the MMD between representations of the source and target domains. $S_{c}$ and $S_{b}$ define the intra-class compactness and the interclass separability.

However, what can we do if there is no class label information in the target domain directly? As we all know, humans can identify unseen classes given only a high-level description. For instance, when provided the description "tall brown animals with long necks", we are able to recognize giraffes. To imitate the ability of humans, \cite{Lampert2009Learning} introduced high-level semantic attributes per class. Assume that ${{a}^{c}}=(a_{1}^{c},...,a_{m}^{c})$ is the attribute representation for class $c$, which has fixed-length binary values with $m$ attributes in all the classes. The classifiers provide estimates of $p(a_m |x)$ for each attribute $a_m$. In the test stage, each target class $y$ obtains its attribute vector $a^y$ in a deterministic way, i.e., $p(a|y)=[\![a=a^y]\!]$. By applying Bayes rule, $p(y|a)=\frac{p(y)}{p({{a}^{y}})}[\![a={{a}^{y}}]\!]$, the posterior of a test class can be calculated as follows:
 \begin{equation}
p(y|x)=\sum\limits_{a\in {{\{0,1\}}^{M}}}{p(y|a)p(a|x)}=\frac{p(y)}{p({{a}^{y}})}\prod\limits_{m=1}^{M}{p(a_{m}^{y}|x)}
\end{equation}

Gebru et al. \cite{Gebru2017Fine} drew inspiration from these works and leveraged attributes to improve performance in the DA of fine-grained recognition. There are multiple independent softmax losses that simultaneously perform attribute and class level to fine-tune the target model. To prevent the independent classifiers from obtaining conflicting labels with attribute and class level, an attribute consistency loss is also implemented.

Occasionally, when fine-tuning the network in unsupervised DA, a label of target data, which is called a pseudo label, can preliminarily be obtained based on the maximum posterior probability. Yan et al. \cite{Yan2017Mind} initialized the target model using the source data and then defined the class posterior probability $p(y_j^t=c|x_j^t)$ by the output of the target model. With $p(y_j^t=c|x_j^t)$, they assigned pseudo-label $\widehat{y_j^t}$ to $x_j^t$ by $\widehat{y_{j}^{t}}=\arg \underset{c}{\mathop{\max }}\,p(y_{j}^{t}=c|x_{j}^{t})$. In \cite{saito2017asymmetric}, two different networks assign pseudo-labels to unlabeled samples, another network is trained by the samples to obtain target discriminative representations. The deep transfer network (DTN) \cite{Zhang2015Deep} used some base classifiers, e.g., SVMs and MLPs, to obtain the pseudo labels for the target samples to estimate the conditional distribution of the target samples and match both the marginal and the conditional distributions with the MMD criterion. When casting the classifier adaptation into the residual learning framework, \cite{Long2016Unsupervised} used the pseudo label to build the conditional entropy $E(\mathcal{D}^t,f^t)$, which ensures that the target classifier $f^t$ fits the target-specific structures well.

\begin{itemize}
\item \textbf{Statistic Criterion}
\end{itemize}

Although some discrepancy-based approaches search for pseudo labels, attribute labels or other substitutes to labeled target data, more work focuses on learning domain-invariant representations via minimizing the domain distribution discrepancy in unsupervised DA.

\begin{figure*}[htbp]
\centering
\label{figure：fig_ref_name}
\includegraphics[width=14cm]{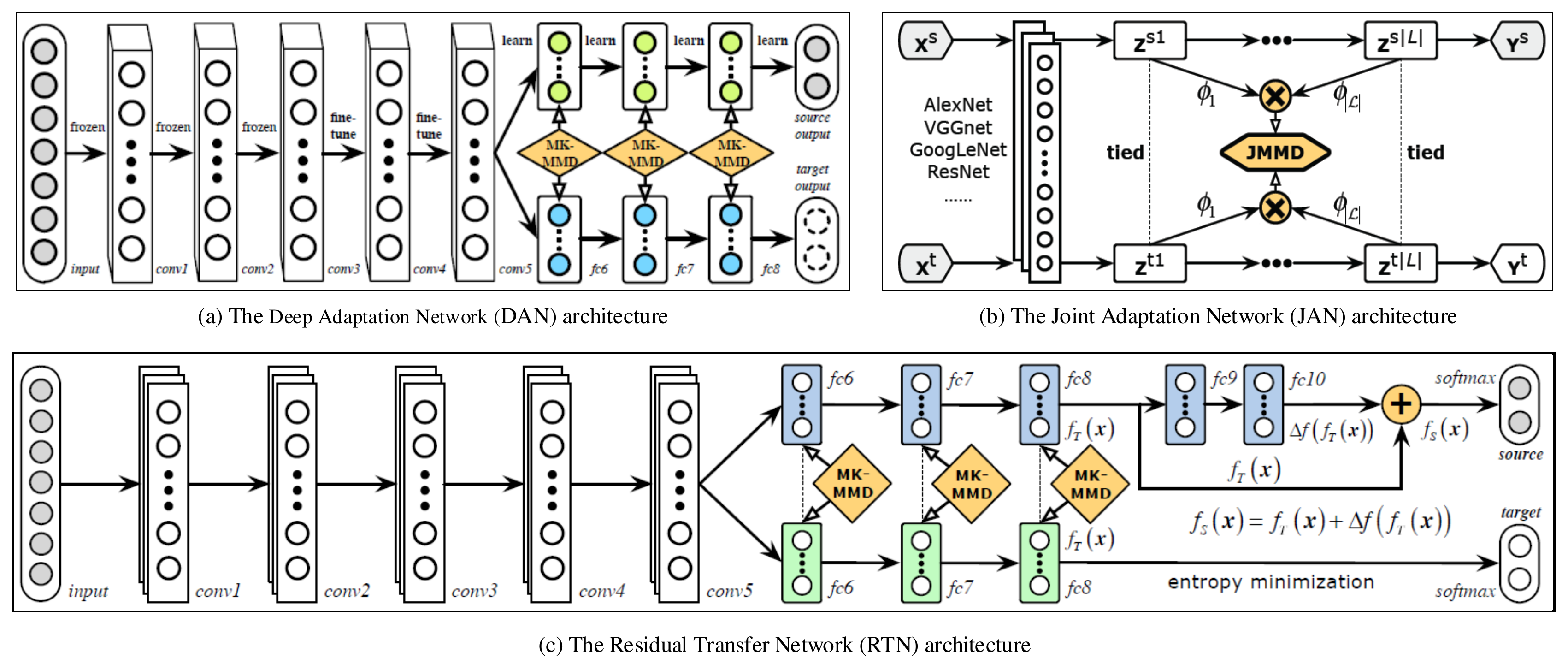}
\caption{Different approaches with the MMD metric. (a) The deep adaptation network (DAN) architecture \cite{Long2015Learning}, (b) the joint adaptation network (JAN) architecture \cite{Long2016Deep} and (c) the residual transfer network (RTN) architecture \cite{Long2016Unsupervised}.}
\end{figure*}

MMD is an effective metric for comparing the distributions between two datasets by a kernel two-sample test \cite{Borgwardt2006Integrating}. Given two distributions $s$ and $t$, the MMD is defined as follows:
\begin{equation}
MM{D^2}(s,t) = \mathop {\sup }\limits_{{{\left\| \phi  \right\|}_{\cal H}} \le 1} \left\| {{E_{{{\rm{x}}^s} \sim s}}[\phi ({{\rm{x}}^s})] - {E_{{{\rm{x}}^t} \sim s}}[\phi ({{\rm{x}}^t})]} \right\|_{\cal H}^2
\end{equation}
where $\phi$ represents the kernel function that maps the original data to a reproducing kernel Hilbert space (RKHS) and ${{\left\| \phi  \right\|}_{\cal H}}\le 1$ defines a set of functions in the unit ball of RKHS $\cal H$.

Based on the above, Ghifary et al. \cite{Ghifary2014Domain} proposed a model that introduced the MMD metric in feedforward neural networks with a single hidden layer. The MMD metric is computed between representations of each domain to reduce the distribution mismatch in the latent space. The empirical estimate of MMD is as follows:
\begin{equation}
MM{D^2}({D_s},{D_t}) = \left\| {\frac{1}{M}\sum\limits_{i = 1}^M {\phi ({\rm{x}}_i^s) - } \frac{1}{N}\sum\limits_{j = 1}^N {\phi ({\rm{x}}_j^t)} } \right\|_H^2
\end{equation}

Subsequently, Tzeng et al. \cite{Tzeng2014Deep} and Long et al. \cite{Long2015Learning} extended MMD to a deep CNN model and achieved great success. The deep domain confusion network (DDC) by Tzeng et al. \cite{Tzeng2014Deep} used two CNNs for the source and target domains with shared weights. The network is optimized for classification loss in the source domain, while domain difference is measured by an adaptation layer with the MMD metric.
\begin{equation}
\mathcal{L}\text{=}{{\mathcal{L}}_{\text{C}}}({{X}^{L}},y)+\lambda MM{{D}^{2}}({{X}^{s}}{{X}^{t}})
\end{equation}
where the hyperparameter $\lambda$ is a penalty parameter. $\mathcal{L}_{C}({{X}^{L}},y)$ denotes classification loss on the available labeled data, $X^L$, and the ground-truth labels, $y$. $MM{{D}^{2}}({{X}^{s}}{{X}^{t}})$ denotes the distance between the source and target data. DDC only adapts one layer of the network, resulting in a reduction in the transferability of multiple layers. Rather than using a single layer and linear MMD, Long et al. \cite{Long2015Learning} proposed the deep adaptation network (DAN) that matches the shift in marginal distributions across domains by adding multiple adaptation layers and exploring multiple kernels, assuming that the conditional distributions remain unchanged. However, this assumption is rather strong in practical applications; in other words, the source classifier cannot be directly used in the target domain. To make it more generalized, a joint adaptation network (JAN) \cite{Long2016Deep} aligns the shift in the joint distributions of input features and output labels in multiple domain-specific layers based on a joint maximum mean discrepancy (JMMD) criterion. \cite{Zhang2015Deep} proposed DTN, where both the marginal and the conditional distributions are matched based on MMD. The shared feature extraction layer learns a subspace to match the marginal distributions of the source and the target samples, and the discrimination layer matches the conditional distributions by classifier transduction. In addition to adapting features using MMD, residual transfer networks (RTNs) \cite{Long2016Unsupervised} added a gated residual layer for classifier adaptation. More recently, \cite{Yan2017Mind} proposed a weighted MMD model that introduces an auxiliary weight for each class in the source domain when the class weights in the target domain are not the same as those in the source domain.

If $\phi$ is a characteristic kernel (i.e., Gaussian kernel or Laplace kernel), MMD will compare all the orders of statistic moments. In contrast to MMD, CORAL \cite{Sun2016Return} learned a linear transformation that aligns the second-order statistics between domains. Sun et al. \cite{Sun2016Deep} extended CORAL to deep neural networks (deep CORAL) with a nonlinear transformation.
\begin{equation}
{{\mathcal{L}}_{CORAL}}\text{=}\frac{1}{4{{d}^{2}}}\left\| {{C}_{S}}-{{C}_{T}} \right\|_{F}^{2}
\end{equation}
where $\|\cdot\|_F^2$ denotes the squared matrix Frobenius norm. $C_S$ and $C_T$ denote the covariance matrices of the source and target data, respectively.

By the Taylor expansion of the Gaussian kernel, MMD can be viewed as minimizing the distance between the weighted sums of all raw moments \cite{Li2015Generative}. The interpretation of MMD as moment matching procedures motivated Zellinger et al. \cite{Zellinger2016Central} to match the higher-order moments of the domain distributions, which we call central moment discrepancy (CMD). An empirical estimate of the CMD metric for the domain discrepancy in the activation space $[a,b]^N$ is given by
\begin{equation}
\begin{aligned}
CM{{D}_{K}}({{X}^{s}},{{X}^{t}})=\frac{1}{(b-a)}{{\left\| E({{X}^{s}})-E({{X}^{t}}) \right\|}_{2}} \\ +\sum\limits_{k=2}^{K}{\frac{1}{{{\left| b-a \right|}^{k}}}}{{\left\| {{C}_{k}}({{X}^{s}})-{{C}_{k}}({{X}^{t}}) \right\|}_{2}}
\end{aligned}
\end{equation}
where ${{C}_{k}}(X)=E({{(x-E(X))}^{k}}$ is the vector of all $k^{th}$-order sample central moments and $E(X)=\frac{1}{\left| X \right|}\sum\nolimits_{x\in X}{x}$ is the empirical expectation.

The association loss ${{\mathcal{L}}_{assoc}}$ proposed by \cite{haeusser2017associative} is an alternative discrepancy measure, it enforces statistical associations between source and target data by making the two-step round-trip probabilities $P^{aba}_{ij}$ be similar to the uniform distribution over the class labels.

\begin{itemize}
\item \textbf{Architecture Criterion}
\end{itemize}

Some other methods optimize the architecture of the network to minimize the distribution discrepancy. This adaptation behavior can be achieved in most deep DA models, such as supervised and unsupervised settings.

Rozantsev et al. \cite{Rozantsev2016Beyond} considered that the weights in corresponding layers are not shared but related by a weight regularizer $r_w(\cdot)$ to account for the differences between the two domains. The weight regularizer $r_w(\cdot)$ can be expressed as the exponential loss function:
\begin{equation}
{{r}_{w}}(\theta _{j}^{s},\theta _{j}^{t})=\exp \left({{\left\| \theta _{j}^{s}-\theta _{j}^{t} \right\|}^{2}}\right)-1
\end{equation}
where $\theta_j^s$ and $\theta_j^t$ denote the parameters of the $j^{th}$ layer of the source and target models, respectively. To further relax this restriction, they allow the weights in one stream to undergo a linear transformation:
\begin{equation}
{{r}_{w}}(\theta _{j}^{s},\theta _{j}^{t})=\exp ({{\left\| {{a}_{j}}\theta _{j}^{s}+{{b}_{j}}-\theta _{j}^{t} \right\|}^{2}})-1
\end{equation}
where $a_j$ and $b_j$ are scalar parameters that encode the linear transformation. The work of Shu et al. \cite{Shu2015Weakly} is similar to \cite{Rozantsev2016Beyond} using weakly parameter-shared layers. The penalty term $\Omega$ controls the relatedness of parameters.
\begin{equation}
\Omega \text{=}\sum\limits_{i=1}^{L}{(\left\| W_{S}^{(l)}-W_{T}^{(l)} \right\|}_{F}^{2}+\left\| b_{S}^{(l)}-b_{T}^{(l)} \right\|_{F}^{2})
\end{equation}
where $\{W_{S}^{(l)},b_{S}^{(l)}\}_{l=1}^{L}$ and $\{W_{T}^{(l)},b_{T}^{(l)}\}_{l=1}^{L}$ are the parameters of the $l^{th}$ layer in the source and target domains, respectively.

\begin{figure}[htbp]
\centering
\label{figure：fig_ref_name}
\includegraphics[width=8cm]{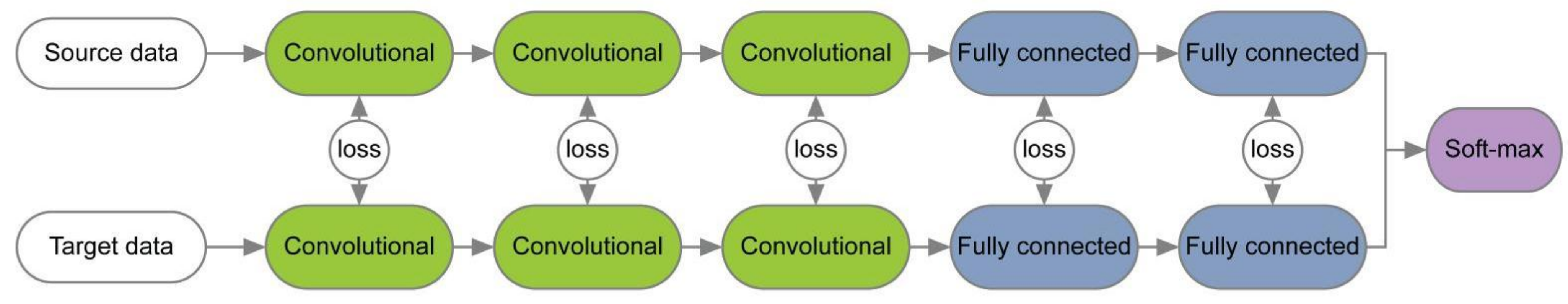}
\caption{The two-stream architecture with related weight \cite{Rozantsev2016Beyond}.}
\end{figure}

Li et al. \cite{Li2016Revisiting} hypothesized that the class-related knowledge is stored in the weight matrix, whereas domain-related knowledge is represented by the statistics of the batch normalization (BN) layer \cite{Ioffe2015Batch}. BN normalizes the mean and standard deviation for each individual feature channel such that each layer receives data from a similar distribution, irrespective of whether it comes from the source or the target domain. Therefore, Li et al. used BN to align the distribution for recomputing the mean and standard deviation in the target domain.
\begin{equation}
BN({{X}^{t}})=\lambda \left(\frac{x-\mu ({{X}^{t}})}{\sigma ({{X}^{t}})}\right)+\beta
\end{equation}
where $\lambda$ and $\beta$ are parameters learned from the target data and $\mu(x)$ and $\sigma(x)$ are the mean and standard deviation computed independently for each feature channel. Based on \cite{Li2016Revisiting}, \cite{carlucci2017autodial} endowed BN layers with a set of alignment parameters which can be learned automatically and can decide the degree of feature alignment required at different levels of the deep network. Furthermore, Ulyanov et al. \cite{Ulyanov2017Improved} found that when replacing BN layers with instance normalization (IN) layers, where $\mu(x)$ and $\sigma(x)$ are computed independently for each channel and each sample, the performance of DA can be further improved.

Occasionally, neurons are not effective for all domains because of the presence of domain biases. For example, when recognizing people, the target domain typically contains one person centered with minimal background clutter, whereas the source dataset contains many people with more clutter. Thus, the neurons that capture the features of other people and clutter are useless. Domain-guided dropout was proposed by \cite{Xiao2016Learning} to solve the problem of multi-DA, and it mutes non-related neurons for each domain. Rather than assigning dropout with a specific dropout rate, it depends on the gain of the loss function of each neuron on the domain sample when the neuron is removed.
\begin{equation}
{{s}_{i}}=\mathcal{L}(g{{(x)}_{\backslash i}})-\mathcal{L}(g(x))\
\end{equation}
where $\mathcal{L}$ is the softmax loss function and $g{{(x)}_{\backslash i}}$ is the feature vector after setting the response of the $i^{th}$ neuron to zero. In \cite{li2017deeper}, each source domain is assigned with different parameters, $\Theta^{(i)}=\Theta^{(0)}+\Delta^{(i)}$, where $\Theta^{(0)}$ is a domain general model, and $\Delta^{(i)}$ is a domain specific bias term. After the low rank parameterized CNNs are trained, $\Theta^{(0)}$ can serve as the classifier for target domain.

\begin{itemize}
\item \textbf{Geometric Criterion}
\end{itemize}

The geometric criterion mitigates the domain shift by integrating intermediate subspaces on a geodesic path from the source to the target domains. A geodesic flow curve is constructed to connect the source and target domains on the Grassmannian. The source and target subspaces are points on a Grassmann manifold. By sampling a fixed \cite{Gopalan2011Domain} or infinite \cite{Grauman2012Geodesic} number of subspaces along the geodesic, we can form the intermediate subspaces to help to find the correlations between domains. Then, both source and target data are projected to the obtained intermediate subspaces to align the distribution.

Inspired by the intermediate representations on the geodesic path, Chopra et al. \cite{chopra2013dlid} proposed a model called deep learning for DA by interpolating between domains (DLID). DLID generates intermediate datasets, starting with all the source data samples and gradually replacing source data with target data. Each dataset is a single point on an interpolating path between the source and target domains. Once intermediate datasets are generated, a deep nonlinear feature extractor using the predictive sparse decomposition is trained in an unsupervised manner.

\subsubsection{Adversarial-Based Approaches}\label{Unsupervised-Adversarial}

Recently, great success has been achieved by the GAN method \cite{Goodfellow2014Generative}, which estimates generative models via an adversarial process. GAN consists of two models: a generative model G that extracts the data distribution and a discriminative model D that distinguishes whether a sample is from G or training datasets by predicting a binary label. The networks are trained on the label prediction loss in a mini-max fashion: simultaneously optimizing G to minimize the loss while also training D to maximize the probability of assigning the correct label:

\begin{equation}
\begin{aligned}
\min_{G}\max_{D}V(D,G)={{\mathbb{E}}_{x\sim {{p}_{data}}(x)}}[\log D(x)] \\ +{{\mathbb{E}}_{z\sim {{p}_{z}}(z)}}[\log (1-D(G(z)))]
\end{aligned}
\end{equation}

In DA, this principle has been employed to ensure that the network cannot distinguish between the source and target domains. \cite{Tzeng2017Adversarial} proposed a unified framework for adversarial-based approaches and summarized the existing approaches according to whether to use a generator, which loss function to employ, or whether to share weights across domains. In this paper, we only categorize the adversarial-based approaches into two subsettings: generative models and non-generative models.

\begin{figure}[htbp]
\centering
\includegraphics[width=8cm]{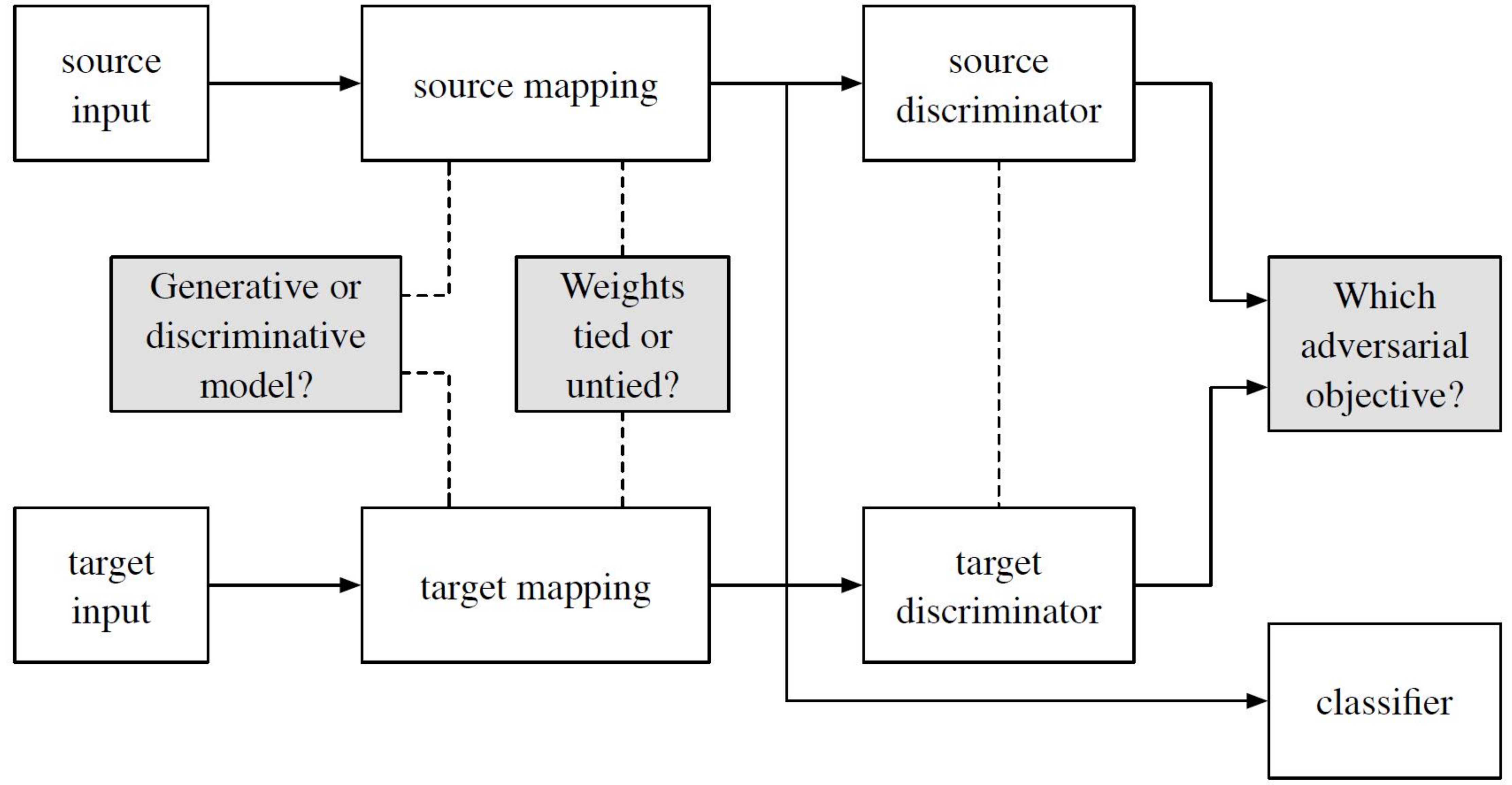}
\caption{Generalized architecture for adversarial domain adaptation. Existing adversarial adaptation methods can be viewed as instantiations of a framework with different choices regarding their properties. \cite{Tzeng2017Adversarial}}
\end{figure}

\begin{itemize}
\item \textbf{Generative Models}
\end{itemize}

Synthetic target data with ground-truth annotations are an appealing alternative to address the problem of a lack of training data. First, with the help of source data, generators render unlimited quantities of synthetic target data, which are paired with synthetic source data to share labels or appear as if they were sampled from the target domain while maintaining labels, or something else. Then, synthetic data with labels are used to train the target model as if no DA were required. Adversarial-based approaches with generative models are able to learn such a transformation in an unsupervised manner based on GAN.

The core idea of CoGAN \cite{Liu2016Coupled} is to generate synthetic target data that are paired with synthetic source ones. It consists of a pair of GANs: $GAN_1$ for generating source data and $GAN_2$ for generating target data. The weights of the first few layers in the generative models and the last few layers in the discriminative models are tied. This weight-sharing constraint allows CoGAN to achieve a domain-invariant feature space without correspondence supervision. A trained CoGAN can adapt the input noise vector to paired images that are from the two distributions and share the labels. Therefore, the shared labels of synthetic target samples can be used to train the target model.

\begin{figure}[htbp]
\centering
\includegraphics[width=8cm]{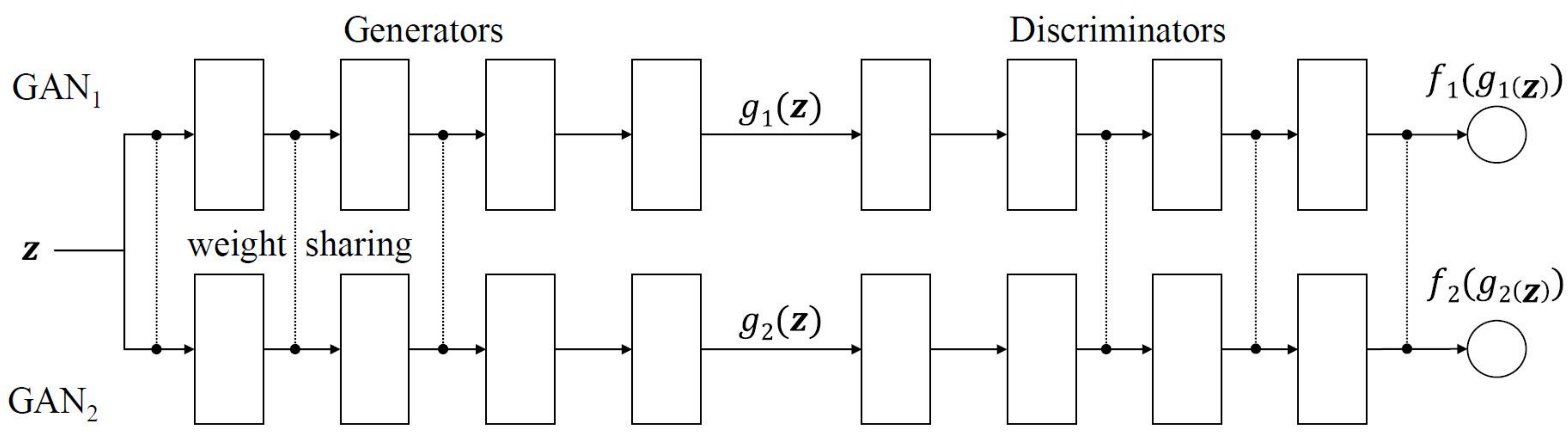}
\caption{The CoGAN architecture. \cite{Liu2016Coupled}}
\end{figure}

More work focuses on generating synthetic data that are similar to the target data while maintaining annotations. Yoo et al. \cite{Yoo2016Pixel} transferred knowledge from the source domain to pixel-level target images with GANs. A domain discriminator ensures the invariance of content to the source domain, and a real/fake discriminator supervises the generator to produce similar images to the target domain. Shrivastava et al. \cite{Shrivastava2016Learning} developed a method for simulated+unsupervised (S+U) learning that uses a combined objective of minimizing an adversarial loss and a self-regularization loss, where the goal is to improve the realism of synthetic images using unlabeled real data. In contrast to other works in which the generator is conditioned only on a noise vector or source images, Bousmalis et al. \cite{Bousmalis2016Unsupervised}  proposed a model that exploits GANs conditioned on both. The classifier T is trained to predict class labels of both source and synthetic images, while the discriminator is trained to predict the domain labels of target and synthetic images. In addition, to expect synthetic images with similar foregrounds and different backgrounds from the same source images, a content similarity is used that penalizes large differences between source and synthetic images for foreground pixels only by a masked pairwise mean squared error \cite{Eigen2014Depth}. The goal of the network is to learn G, D and T by solving the optimization problem:

\begin{equation}
\begin{aligned}
\min_{G,T}\max_{D}V(D,G)=\alpha {{\mathcal{L}}_{d}}(D,G)\\+\beta {{\mathcal{L}}_{t}}(T,G)+\gamma {{\mathcal{L}}_{c}}(G)
\end{aligned}
\end{equation}
where $\alpha$, $\beta$, and $\gamma$ are parameters that control the trade-off between the losses. ${{\mathcal{L}}_{d}}$, ${{\mathcal{L}}_{t}}$ and ${{\mathcal{L}}_{c}}$ are the adversarial loss, softmax loss and content-similarity loss, respectively.

\begin{figure}[htbp]
\centering
\includegraphics[width=8cm]{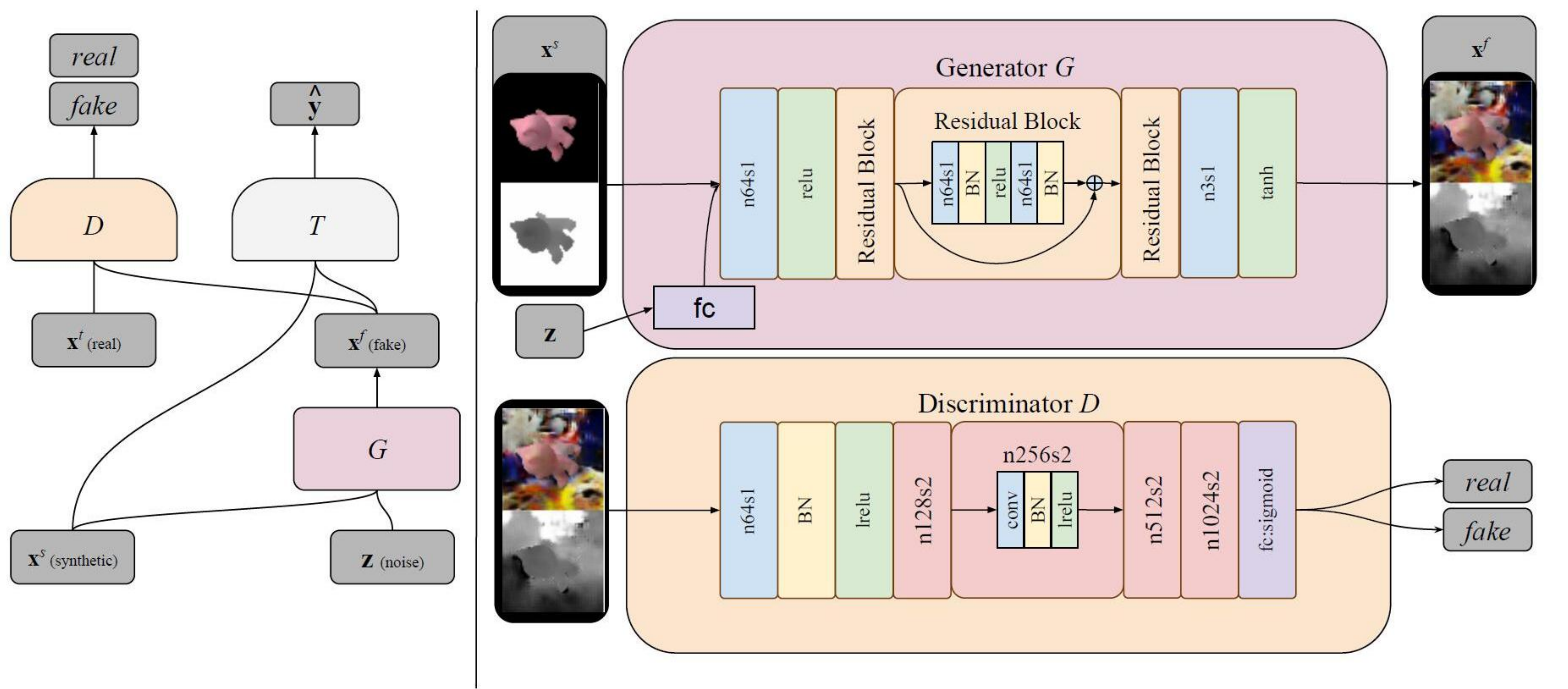}
\caption{The model that exploits GANs conditioned on noise vector and source images. \cite{Bousmalis2016Unsupervised}}
\end{figure}

\begin{itemize}
\item \textbf{Non-Generative Models}
\end{itemize}

The key of deep DA is learning domain-invariant representations from source and target samples. With these representations, the distribution of both domains can be similar enough such that the classifier is fooled and can be directly used in the target domain even if it is trained on source samples. Therefore, whether the representations are domain-confused or not is crucial to transferring knowledge. Inspired by GAN, domain confusion loss, which is produced by the discriminator, is introduced to improve the performance of deep DA without generators.

\begin{figure}[htbp]
\centering
\includegraphics[width=8cm]{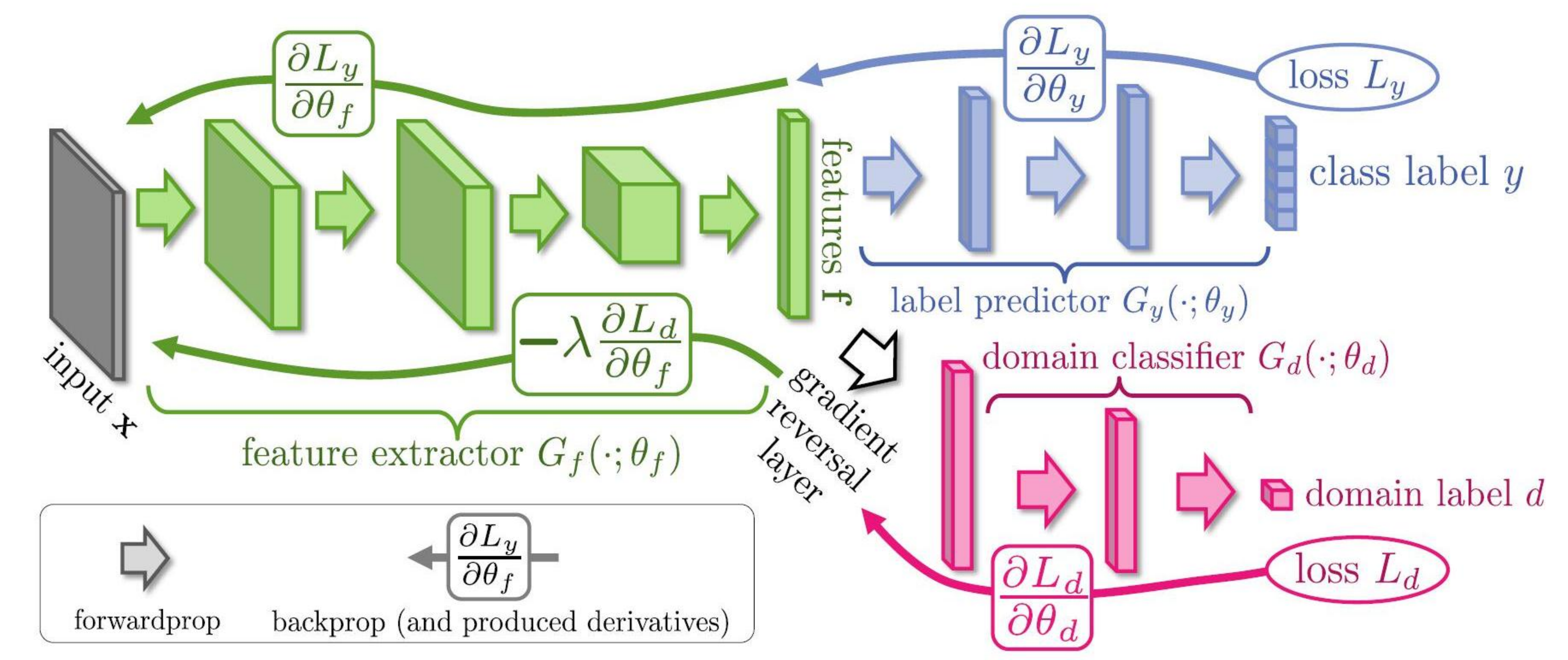}
\caption{The domain-adversarial neural network (DANN) architecture. \cite{Ganin2015Unsupervised}}
\end{figure}

The domain-adversarial neural network (DANN) \cite{Ganin2015Unsupervised} integrates a gradient reversal layer (GRL) into the standard architecture to ensure that the feature distributions over the two domains are made similar. The network consists of shared feature extraction layers and two classifiers. DANN minimizes the domain confusion loss (for all samples) and label prediction loss (for source samples) while maximizing domain confusion loss via the use of the GRL. In contrast to the above methods, the adversarial discriminative domain adaptation (ADDA) \cite{Tzeng2017Adversarial} considers independent source and target mappings by untying the weights, and the parameters of the target model are initialized by the pre-trained source one. This is more flexible because of allowing more domain-specific feature extractions to be learned. ADDA minimizes the source and target representation distances through iteratively minimizing these following functions, which is most similar to the original GAN:

\begin{equation*}
\begin{split}
\min_{{{M}^{s}},C}&{{\mathcal{L}}_{cls}}({{X}^{s}},{{Y}^{s}})=\\
&-{{\mathbb{E}}_{({{x}^{s}},{{y}^{s}})\sim ({{X}^{s}},{{Y}^{s}})}}\sum\limits_{k=1}^{K}{{\mathbbm{1}_{\left[ k={{y}^{s}} \right]}}}\log C({{M}^{s}}({{x}^{s}}))
\end{split}
\end{equation*}
\begin{equation*}
\begin{split}
\min_{D}{{\mathcal{L}}_{advD}}({{X}^{s}},&{{X}^{t}},{{M}^{s}},{{M}^{t}})=\\
&-{{\mathbb{E}}_{({{x}^{s}})\sim ({{X}^{s}})}}[\log D({{M}^{s}}({{x}^{s}}))]\\
&-{{\mathbb{E}}_{({{x}^{t}})\sim ({{X}^{t}})}}[\log (1-D({{M}^{t}}({{x}^{t}})))]
\end{split}
\end{equation*}
\begin{equation}
\begin{split}
\min_{{{M}^{s}},{{M}^{t}}}{{\mathcal{L}}_{advM}}&({{M}^{s}},{{M}^{t}})=\\
&-{{\mathbb{E}}_{({{x}^{t}})\sim ({{X}^{t}})}}[\log D({{M}^{t}}({{x}^{t}}))]
\end{split}
\end{equation}
where the mappings ${{M}^{s}}$ and ${{M}^{t}}$ are learned from the source and target data, ${{X}^{s}}$ and ${{X}^{t}}$. $C$ represents a classifier working on the source domain.  The first classification loss function ${{\mathcal{L}}_{cls}}$ is optimized by training the source model using the labeled source data. The second function ${{\mathcal{L}}_{advD}}$ is minimized to train the discriminator, while the third function ${{\mathcal{L}}_{advM}}$ is learning a representation that is domain invariant.

\begin{figure}[htbp]
\centering
\includegraphics[width=8cm]{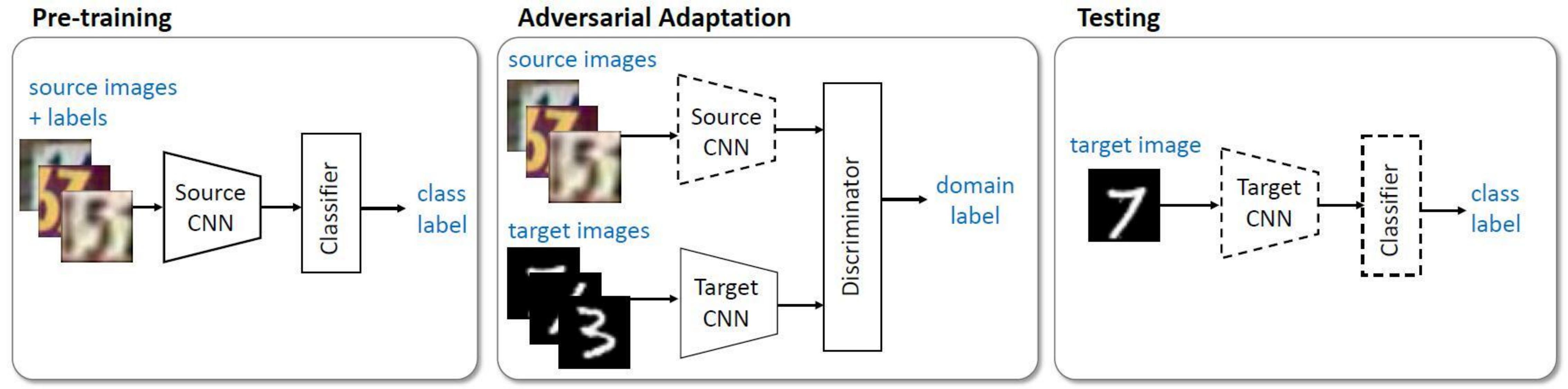}
\caption{The Adversarial discriminative domain adaptation (ADDA) architecture. \cite{Tzeng2017Adversarial}}
\end{figure}

Tzeng et al. \cite{Tzeng2015Simultaneous} proposed adding an additional domain classification layer that performs binary domain classification and designed a domain confusion loss to encourage its prediction to be as close as possible to a uniform distribution over binary labels. Unlike previous methods that match the entire source and target domains, Cao et al. introduced a selective adversarial network (SAN) \cite{Cao2017Partial} to address partial transfer learning from large domains to small domains, which assumes that the target label space is a subspace of the source label space. It simultaneously avoids negative transfer by filtering out outlier source classes, and it promotes positive transfer by matching the data distributions in the shared label space via splitting the domain discriminator into many class-wise domain discriminators. \cite{motiian2017few} encoded domain labels and class labels to produce four groups of pairs, and replaced the typical binary adversarial discriminator by a four-class discriminator. Volpi et al. \cite{volpi2017adversarial} trained a feature generator (S) to perform data augmentation in the source feature space and obtained a domain invariant feature through playing a minimax game against features from S.

Rather than using discriminator to classify domain label, some papers make some other explorations. Inspired by Wasserstein GAN \cite{arjovsky2017wasserstein}, Shen et al. \cite{Shen2017Wasserstein} utilized discriminator to estimate empirical Wasserstein distance between the source and target samples and optimized the feature extractor network to minimize the distance in an adversarial manner. In \cite{saito2017maximum}, two classifiers are treated as discriminators and are trained to maximize the discrepancy to detect target samples outside the support of the source, while a feature extractor is trained to minimize the discrepancy by generating target features near the support.

\subsubsection{Reconstruction-Based Approaches}\label{Unsupervised-Reconstruction}

In DA, the data reconstruction of source or target samples is an auxiliary task that simultaneously focuses on creating a shared representation between the two domains and keeping the individual characteristics of each domain.

\begin{itemize}
\item \textbf{Encoder-Decoder Reconstruction}
\end{itemize}

The basic autoencoder framework \cite{Bengio2009Learning} is a feedforward neural network that includes the encoding and decoding processes. The autoencoder first encodes an input to some hidden representation, and then it decodes this hidden representation back to a reconstructed version. The DA approaches based on encoder-decoder reconstruction typically learn the domain-invariant representation by a shared encoder and maintain the domain-special representation by a reconstruction loss in the source and target domains.

Xavier and Bengio \cite{Glorot2011Domain} proposed extracting a high-level representation based on stacked denoising autoencoders (SDA) \cite{Vincent2010Stacked}.  By reconstructing the union of data from various domains with the same network, the high-level representations can represent both the source and target domain data. Thus, a linear classifier that is trained on the labeled data of the source domain can make predictions on the target domain data with these representations. Despite their remarkable results, SDAs are
limited by their high computational cost and lack of scalability to high-dimensional features. To address these crucial limitations, Chen et al. \cite{Chen2012Marginalized} proposed the marginalized SDA (mSDA), which marginalizes noise with linear denoisers; thus, parameters can be computed in closed-form and do not require stochastic gradient descent.

The deep reconstruction classification network (DRCN) proposed in \cite{Ghifary2016Deep} learns a shared encoding representation that provides useful information for cross-domain object recognition. DRCN is a CNN architecture that combines two pipelines with a shared encoder. After a representation is provided by the encoder, the first pipeline, which is a CNN, works for supervised classification with source labels, whereas the second pipeline, which is a deconvolutional network, optimizes for unsupervised reconstruction with target data.
\begin{equation}
\min \lambda {{\mathcal{L}}_{c}}(\{{{\theta }_{enc}},{{\theta }_{lab}}\})+(1-\lambda ){{\mathcal{L}}_{r}}(\{{{\theta }_{enc}},{{\theta }_{dec}}\})
\end{equation}
where $\lambda$ is a hyper-parameter that controls the trade-off between classification and reconstruction. ${\theta }_{enc}$, ${\theta }_{dec}$ and ${\theta }_{lab}$ denote the parameters of the encoder, decoder and source classifier, respectively. ${\mathcal{L}}_{c}$ is cross-entropy loss for classification, and ${\mathcal{L}}_{r}$ is squared loss ${\parallel x-f_r(x) \parallel}_2^2$ for reconstruction in which $f_r(x)$ is the reconstruction of $x$.

\begin{figure}[htbp]
\centering
\includegraphics[width=8cm]{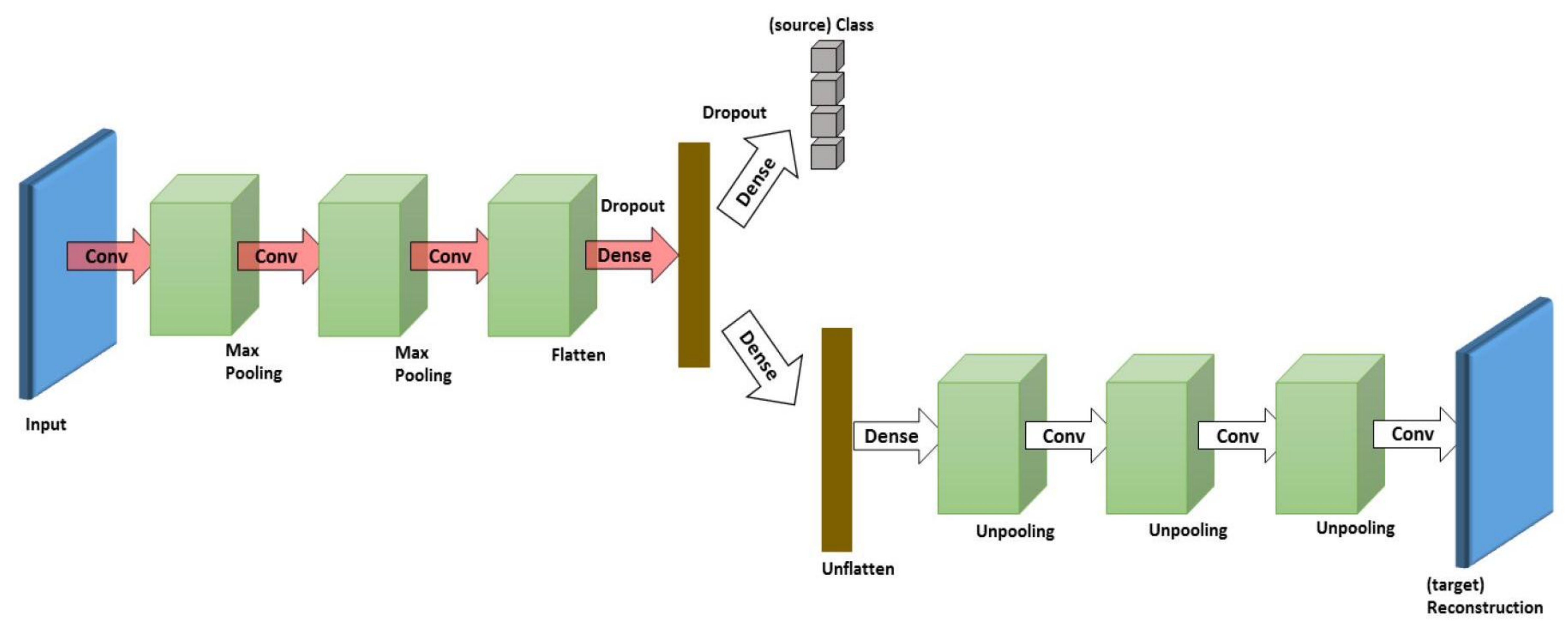}
\caption{The deep reconstruction classification network (DRCN) architecture. \cite{Ghifary2016Deep}}
\end{figure}

Domain separation networks (DSNs) \cite{Bousmalis2016Domain} explicitly and jointly model both private and shared components of the domain representations. A shared-weight encoder learns to capture shared representations, while a private encoder is used for domain-specific components in each domain. Additionally, a shared decoder learns to reconstruct the input samples by both the private and shared representations. Then, a classifier is trained on the shared representation. By partitioning the space in such a manner, the shared representations will not be influenced by domain-specific representations such that a better transfer ability can be obtained. Finding that the separation loss is simple and that the private features are only used for reconstruction in DSNs, \cite{tsai2017adversarial} reinforced them by incorporating a hybrid adversarial learning in a separation network and an adaptation network.

Zhuang et al. \cite{Zhuang2015Supervised} proposed transfer learning with deep autoencoders (TLDA), which consists of two encoding layers. The distance in distributions between domains is minimized with KL divergence in the embedding encoding layer, and label information of the source domain is encoded using a softmax loss in the label encoding layer. Ghifary et al. \cite{Ghifary2015Domain} extended the autoencoder into a model that jointly learns two types of data-reconstruction tasks taken from related domains: one is self-domain reconstruction, and the other is between-domain reconstruction.

\begin{itemize}
\item \textbf{Adversarial Reconstruction}
\end{itemize}

Dual learning was first proposed by Xia et al. \cite{Xia2016Dual} to reduce the requirement of labeled data in natural language processing. Dual learning trains two "opposite" language translators, e.g., A-to-B and B-to-A. The two translators represent a primal-dual pair that evaluates how likely the translated sentences belong to the targeted language, and the closed loop measures the disparity between the reconstructed and the original ones. Inspired by dual learning, adversarial reconstruction is adopted in deep DA with the help of dual GANs.

Zhu et al. \cite{Zhu2017Unpaired} proposed a cycle GAN that can translate the characteristics of one image domain into the other in the absence of any paired training examples. Compared to dual learning, cycle GAN uses two generators rather than translators, which learn a mapping $G:X\rightarrow Y$ and an inverse mapping $F:Y\rightarrow X$. Two discriminators, $D_X$ and $D_Y$, measure how realistic the generated image is ($G(X)\approx Y$ or $G(Y)\approx X$) by an adversarial loss  and how well the original input is reconstructed after a sequence of two generations ($F(G(X))\approx X$ or $G(F(Y))\approx Y$) by a cycle consistency loss (reconstruction loss). Thus, the distribution of images from $G(X)$ (or $F(Y)$) is indistinguishable from the distribution $Y$ (or $X$).
\begin{equation*}
\begin{split}
{{\mathcal{L}}_{GAN}}(G,{{D}_{Y}},X,Y)={{\mathbb{E}}_{y\sim {{p}_{data}}(y)}}[\log {{D}_{Y}}(y)]\\+{{\mathbb{E}}_{x\sim {{p}_{data}}(x)}}[\log (1-{{D}_{Y}}(G(x)))]
\end{split}
\end{equation*}
\begin{equation}
\begin{split}
{{\mathcal{L}}_{cyc}}(G,F)={{\mathbb{E}}_{x\sim data(x)}}[{{\left\| F(G(x))-x \right\|}_{1}}]\\+{{\mathbb{E}}_{y\sim data(y)}}[{{\left\| G(F(y))-y \right\|}_{1}}]
\end{split}
\end{equation}
where ${\mathcal{L}}_{GAN}$ is the adversarial loss produced by discriminator $D_Y$ with mapping function $G:X\rightarrow Y$. ${\mathcal{L}}_{cyc}$ is the reconstruction loss using L1 norm.

\begin{figure}[htbp]
\centering
\includegraphics[width=8cm]{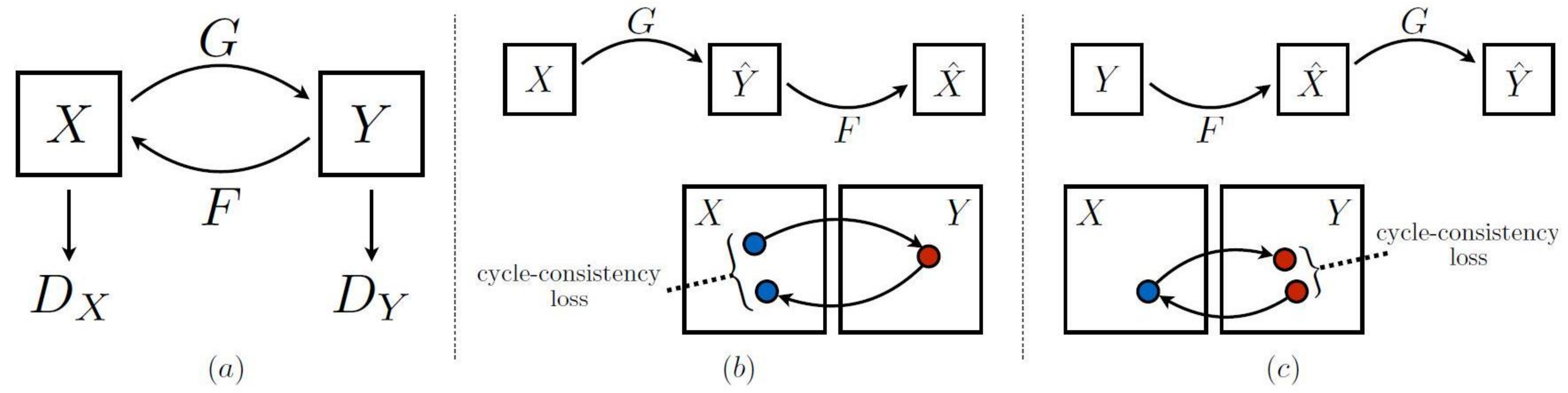}
\caption{The cycle GAN architecture. \cite{Zhu2017Unpaired}}
\end{figure}

The dual GAN \cite{Yi2017DualGAN} and the disco GAN \cite{Kim2017Learning} were proposed at the same time, where the core idea is similar to cycle GAN. In dual GAN, the generator is configured with skip connections between mirrored downsampling and upsampling layers \cite{Ronneberger2015U,Isola2016Image}, making it a U-shaped net to share low-level information (e.g., object shapes, textures, clutter, and so forth). For discriminators, the Markovian patch-GAN \cite{Li2016Precomputed} architecture is employed to capture local high-frequency information. In disco GAN, various forms of distance functions, such as mean-square error (MSE), cosine distance, and hinge loss, can be used as the reconstruction loss, and the network is applied to translate images, changing specified attributes including hair color, gender and orientation while maintaining all other components.%Editor: Please ensure that the intended meaning has been maintained in the above edit.

\subsubsection{Hybrid Approaches}

To obtain better performance, some of the aforementioned methods have been used simultaneously. \cite{Tzeng2015Simultaneous} combined a domain confusion loss and a soft label loss, while \cite{Long2016Unsupervised} used both statistic (MMD) and architecture criteria (adapt classifier by residual function) for unsupervised DA. \cite{Yan2017Mind} introduced class-specific auxiliary weights assigned by the pseudo-labels into the original MMD. In DSNs \cite{Bousmalis2016Domain}, encoder-decoder reconstruction approaches separate representations into private and shared representations, while the MMD criterion or domain confusion loss is helpful to make the shared representations similar and soft subspace orthogonality constraints ensure dissimilarity between the private and shared representations. \cite{Rozantsev2016Beyond} used the MMD between the learned source and target representations and also allowed the weights of the corresponding layers to differ. \cite{Zhuang2015Supervised} learned domain-invariant representations by encoder-decoder reconstruction approaches and the KL divergence.

\subsection{Heterogeneous Domain Adaptation}

In heterogeneous DA, the feature spaces of the source and target domains are not the same, $Xs \neq Xt$, and the dimensions of the feature spaces may also differ. According to the divergence of feature spaces, heterogeneous DA can be further divided into two scenarios. In one scenario, the source and target domain both contain images, and the divergence of feature spaces is mainly caused by different sensory devices (e.g., visual light (VIS) vs. near-infrared (NIR) or RGB vs. depth) and different styles of images (e.g., sketches vs. photos). In the other scenario, there are different types of media in source and target domain (e.g., text vs. image and language vs. image). Obviously, the cross-domain gap of the second scenario is much larger.

Most heterogeneous DA with shallow methods fall into two categories: symmetric transformation and asymmetric transformation. The symmetric transformation learns feature transformations to project the source and target features onto a common subspace. Heterogeneous feature augmentation (HFA) \cite{duan2012learning} first transformed the source and target data into a common subspace using projection matrices $P$ and $Q$ respectively, then proposed two new feature mapping functions, $\varphi _{s}\left ( x^{s} \right )=\left [ Px^{s}, x^{s}, 0_{d_{t}} \right ]^{T}$ and $\varphi _{t}\left ( x^{t} \right )=\left [ Qx^{t}, 0_{d_{s}}, x^{t} \right ]^{T}$, to augment the transformed data with their original features and zeros. These projection matrices are found using standard SVM with hinge loss in both the linear and nonlinear cases and an alternating optimization algorithm is proposed to simultaneously solve the dual SVM and to find the optimal transformations. \cite{wang2011heterogeneous} treated each input domain as a manifold which is represented by a Laplacian matrix, and used labels rather than correspondences to align the manifolds. The asymmetric transformation transforms one of source and target features to align with the other. \cite{zhou2014heterogeneous} proposed a sparse and class-invariant feature transformation matrix to map the weight vector of classifiers learned from the source domain to the target domain. The asymmetric regularized cross-domain transfer (ARC-t) \cite{kulis2011you} used asymmetric, non-linear transformations learned in Gaussian RBF kernel space to map the target data to the source domain. Extended from \cite{saenko2010adapting}, ARC-t performed asymmetric transformation based on metric learning, and transfer knowledge between domains with different dimensions through changes of the regularizer. Since we focus on deep DA, we refer the interested readers to \cite{Day2017A}, which summarizes shallow approaches of heterogeneous DA.

However, as for deep methods, there is not much work focused on heterogeneous DA so far. The special and effective methods of heterogeneous deep DA have not been proposed, and heterogeneous deep DA is still performed similar to some approaches of homogeneous DA.

\subsubsection{Discrepancy-Based Approach}

In discrepancy-based approaches, the network generally shares or reuses the first n layers between the source and target domains, which limits the feature spaces of the input to the same dimension. However, in heterogeneous DA, the dimensions of the feature spaces of source domain may differ from those of target domain.

In first scenario of heterogeneous DA, the images in different domains can be directly resized into the same dimensions, so the Class Criterion and Statistic Criterion are still effective and are mainly used. For example, given an RGB image and its paired depth image, \cite{Gupta2016Cross} used the mid-level representation learned by CNNs as a supervisory signal to re-train a CNN on depth images. To transform an RGB object detector into a RGB-D detector without needing complete RGB-D data, Hoffman et al. \cite{Hoffman2016Cross} first trained an RGB network using labeled RGB data from all categories and finetuned the network with labeled depth data from partial categories, then combined mid-level RGB and depth representations at fc6 to incorporate both modalities into the final object class prediction. \cite{mittal2015composite} first trained the network using large face database of photos and then finetuned it using small database of composite sketches; \cite{liu2016transferring} transferred the VIS deep networks to the NIR domain in the same way.

In second scenario, the features of different media can not be directly resized into the same dimensions. Therefore, discrepancy-based methods fail to work without extra process. \cite{Shu2015Weakly} proposed weakly shared DTNs to transfer labeled information across heterogeneous domains, particularly from the text domain to the image domain. DTNs take paired data, such as text and image, as input to two SAEs, followed by weakly parameter-shared network layers at the top. Chen et al. \cite{Chen2016Transfer} proposed transfer neural trees (TNTs), which consist of two stream networks to learn a domain-invariant feature representation for each modality. Then, a transfer neural decision forest (Transfer-NDF) \cite{Kontschieder2014Neural,Kontschieder2016Deep} is used with stochastic pruning for adapting representative neurons in the prediction layer.

\subsubsection{Adversarial-Based Approach}

Using Generative Models can generate the heterogeneous target data while transferring some information of source domain to them. \cite{Taigman2016Unsupervised} employed a compound loss function that consists of a multiclass GAN loss, a regularizing component and an f-constancy component to transfer unlabeled face photos to emoji images. To generate images for birds and flowers based on text, \cite{reed2016generative} trained a GAN conditioned on text features encoded by a hybrid character-level convolutional-recurrent neural network. \cite{zhang2017stackgan} proposed stacked generative adversarial networks (StackGAN) with conditioning augmentation for synthesizing photo-realistic images from text. It decomposes the synthesis problem into several sketch-refinement processes. Stage-I GAN sketches the primitive shape and basic colors of the object to yield low-resolution image, and Stage-II GAN completes details of the object to produce a high-resolution photo-realistic image.

\begin{figure}[htbp]
\centering
\includegraphics[width=8cm]{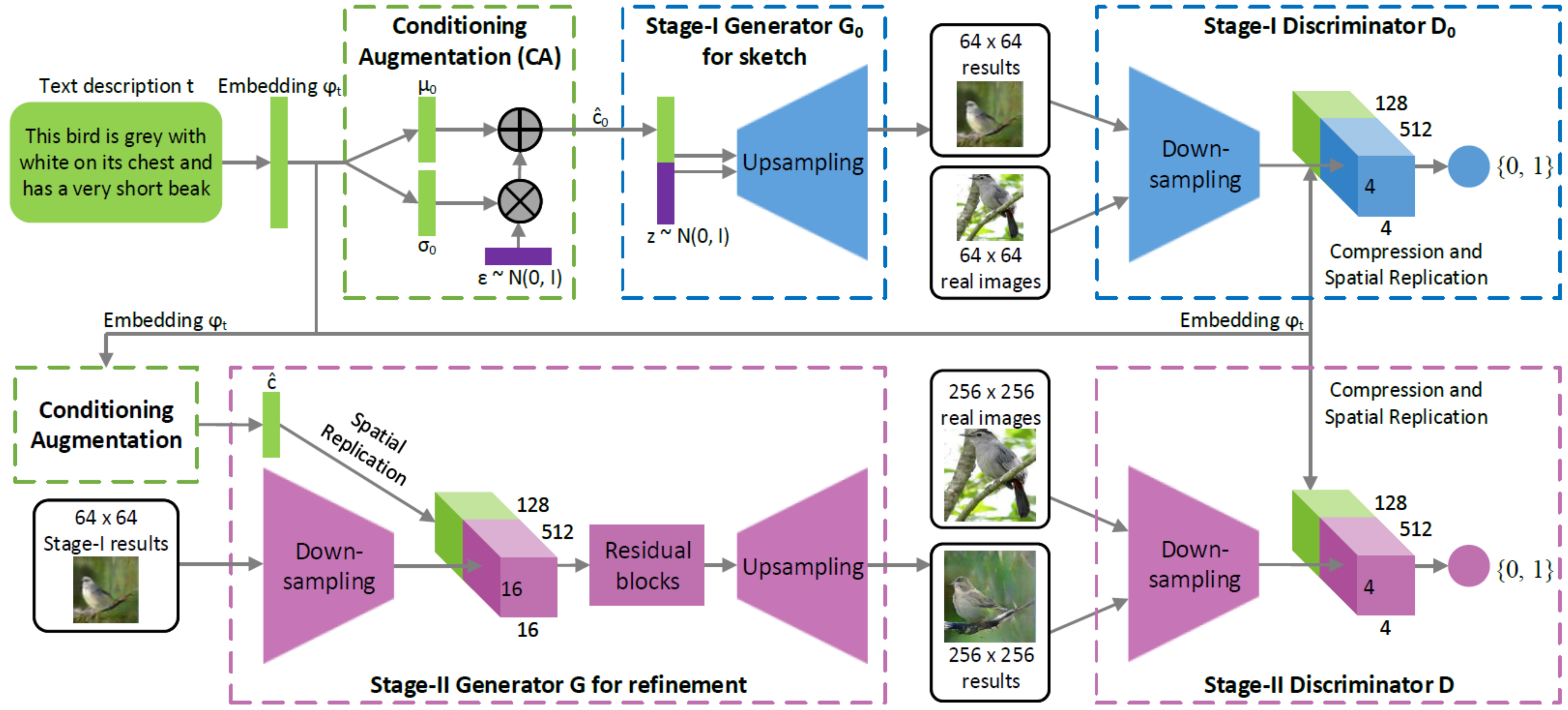}
\caption{The StackGAN architecture. \cite{zhang2017stackgan}}
\end{figure}

\subsubsection{Reconstruction-Based Approach}

The Adversarial Reconstruction can be used in heterogeneous DA as well. For example, the cycle GAN \cite{Zhu2017Unpaired}, dual GAN \cite{Yi2017DualGAN} and disco GAN \cite{Kim2017Learning} used two generators, $G_A$ and $G_B$, to generate sketches from photos and photos from sketches, respectively. Based on cycle GAN \cite{Zhu2017Unpaired}, \cite{wang2017high} proposed a multi-adversarial network to avoid artifacts of facial photo-sketch synthesis by leveraging the implicit presence of feature maps of different resolutions in the generator subnetwork.

\section{Multi-Step Domain Adaptation}

For multi-step DA, the selection of the intermediate domain is problem specific, and different problems may have different strategies.

\subsection{Hand-Crafted Approaches}

Occasionally, the intermediate domain can be selected by experience, that is, it is decided in advance. For example, when the source domain is image data and the target domain is composed of text data, some annotated images will clearly be crawled as intermediate domain data.

With the common sense that nighttime light intensities can be used as a proxy for economic activity, Xie et al. \cite{Xie2015Transfer} transferred knowledge from daytime satellite imagery to poverty prediction with the help of some nighttime light intensity information as an intermediate domain.

\subsection{Instance-Based Approaches}

In other problems where there are many candidate intermediate domains, some automatic selection criterion should be considered. Similar to the instance-transfer approaches proposed by Pan \cite{Pan2010A}, because the samples of the source domain cannot be used directly, the mixture of certain parts of the source and target data can be useful for constructing the intermediate domain.

Tan et al. \cite{tan2017distant} proposed distant domain transfer learning (DDTL), where long-distance domains fail to transfer knowledge by only one intermediate domain but can be related via multiple intermediate domains. DDTL gradually selects unlabeled data from the intermediate domains by minimizing reconstruction errors on the selected instances in the source and intermediate domains and all the instances in the target domain simultaneously. With removal of the unrelated source data, the selected intermediate domains gradually become closer  to the target domain from the source domain:
\begin{equation}
\begin{split}
{{\mathcal{J}}_{1}}({{f}_{e}},{{f}_{d}},{{v}_{S}},{{v}_{T}})&=\frac{1}{{{n}_{S}}}\sum\limits_{i=1}^{{{n}_{S}}}{v_{S}^{i}}\left\| \hat{x}_{S}^{i}-x_{S}^{i} \right\|_{2}^{2}\\&+\frac{1}{{{n}_{I}}}\sum\limits_{i=1}^{{{n}_{I}}}{v_{I}^{i}}\left\| \hat{x}_{I}^{i}-x_{I}^{i} \right\|_{2}^{2}\\&+\frac{1}{{{n}_{T}}}\sum\limits_{i=1}^{{{n}_{T}}}{\left\| \hat{x}_{T}^{i}-x_{T}^{i} \right\|_{2}^{2}}+R({{v}_{S}},{{v}_{T}})
\end{split}
\end{equation}
where $\hat{x}_{S}^{i}$, $\hat{x}_{T}^{i}$ and $\hat{x}_{I}^{i}$ are reconstructions of source data ${S}^{i}$, target data ${T}^{i}$ and intermediate data ${I}^{i}$ based on the autoencoder, respectively, and $f_e$ and $f_d$ are the parameters of the encoder and decoder, respectively. ${{v}_{S}}={{(v_{S}^{1},...,v_{S}^{{{n}_{S}}})}^\top}$ and ${{v}_{I}}={{(v_{I}^{1},...,v_{I}^{{{n}_{I}}})}^\top}$, $v_{S}^{i}$, $v_{I}^{i}\in{0,1}$ are selection indicators for the $i^{th}$ source and intermediate instance, respectively. $R({{v}_{S}},{{v}_{T}})$ is a regularization term that avoids all values of $v_{S}$ and $v_{I}$ being zero.

The DLID model \cite{chopra2013dlid} mentioned in Section \ref{Discrepancy-Supervised} (Geometric Criterion) constructs the intermediate domains with a subset of the source and target domains, where source samples are gradually replaced by target samples.

\subsection{Representation-Based Approaches}

Representation-based approaches freeze the previously trained network and use their intermediate representations as input to the new network. Rusu et al. \cite{Rusu2016Progressive} introduced progressive networks that have the ability to accumulate and transfer knowledge to new domains over a sequence of experiences. To avoid the target model losing its ability to solve the source domain, they constructed a new neural network for each domain, while transfer is enabled via lateral connections to features of previously learned networks. In the process, the parameters in the latest network are frozen to remember knowledge of intermediate domains.

\begin{figure}[htbp]
\centering
\includegraphics[width=5cm]{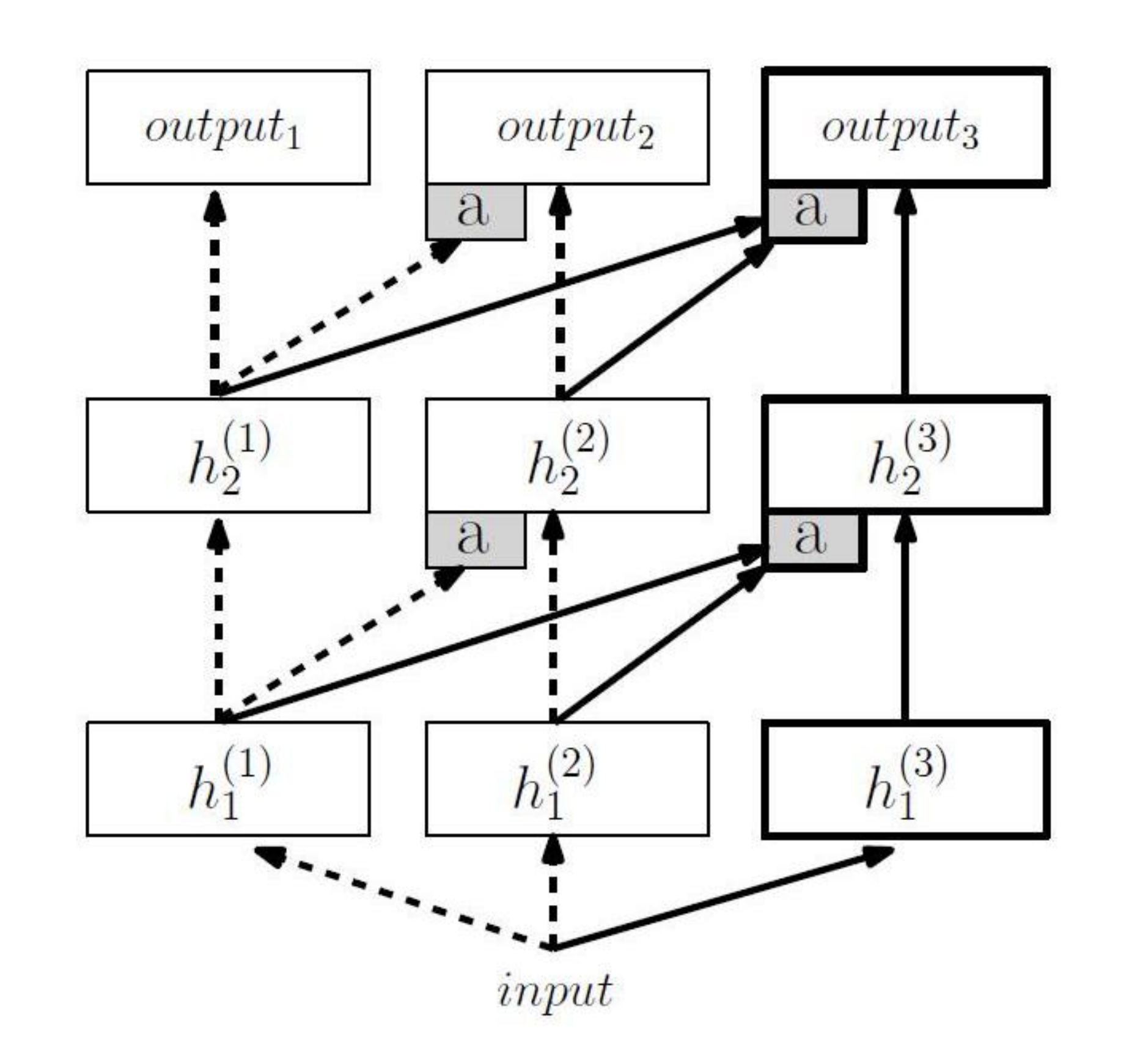}
\caption{The progressive network architecture. \cite{Rusu2016Progressive}}
\end{figure}

\section{Application of Deep Domain Adaptation}

Deep DA techniques have recently been successfully applied in many real-world applications, including image classification, object recognition, face recognition, object detection, style translation, and so forth. In this section, we present different application examples using various visual deep DA methods. Because the information of commonly used datasets for evaluating the performance is provided in \cite{Zhang2017Transfer} in detail, we do not introduce it in this paper.

\subsection{Image Classification}

\begin{table*}[htbp]
\centering
\caption{Comparison between Transfer Learning and Non-Adaptation Learning Methods}
\begin{tabular}{|c|c|c|c|c|c|c|c|}
	\hline
    \tabincell{l}{Data Set \\(reference)} & Source vs. Target & Baselines & \multicolumn{5}{|c|}{Deep Domain Adaptation Methods}\\ \hline
	 & & \textbf{AlexNet} & \textbf{DDC} & \textbf{DAN} & \textbf{RTN} & \textbf{JAN} & \textbf{DANN} \\ \cline{2-8}
     & A vs. W	& 61.6$\pm$0.5 & 61.8$\pm$0.4 & 68.5         & 73.3$\pm$0.3 & 75.2$\pm$0.4 & 73.0$\pm$0.5 \\ \cline{2-8}
     & D vs. W	& 95.4$\pm$0.3 & 95.0$\pm$0.5 & 96.0$\pm$0.3 & 96.8$\pm$0.2 & 96.6$\pm$0.2 & 96.4$\pm$0.3 \\ \cline{2-8}
     Office-31 Dataset & W vs. D & 99.0$\pm$0.2 & 98.5$\pm$0.4 & 99.0$\pm$0.3 & 99.6$\pm$0.1 & 99.6$\pm$0.1 & 99.2$\pm$0.3 \\ \cline{2-8}
     ACC (unit:\%)\cite{Long2016Deep}& A vs. D	& 63.8$\pm$0.5 & 64.4$\pm$0.3 & 67.0$\pm$0.4 & 71.0$\pm$0.2	& 72.8$\pm$0.3 & 72.3$\pm$0.3 \\ \cline{2-8}
     & D vs. A	& 51.1$\pm$0.6 & 52.1$\pm$0.6 & 54.0$\pm$0.5 & 50.5$\pm$0.3	& 57.5$\pm$0.2 & 53.4$\pm$0.4 \\ \cline{2-8}
     & W vs. A	& 49.8$\pm$0.4 & 52.2$\pm$0.4 & 53.1$\pm$0.5 & 51.0$\pm$0.1	& 56.3$\pm$0.2 & 51.2$\pm$0.5 \\ \cline{2-8}
     & Avg & 70.1 & 70.6 & 72.9 & 73.7 & 76.3 & 74.3 \\ \hline

     & & \textbf{AlexNet} & \textbf{Deep CORAL} & \textbf{CMD} & \textbf{DLID} & \textbf{AdaBN} &\textbf{DANN} \\ \cline{2-8}
     & A vs. W	& 61.6 & 66.4 & 77.0$\pm$0.6 & 51.9 & 74.2 & 73 \\ \cline{2-8}
     & D vs. W	& 95.4 & 95.7 & 96.3$\pm$0.4 & 78.2 & 95.7 & 96.4 \\ \cline{2-8}
     Office-31 Dataset & W vs. D	& 99.0 & 99.2 & 99.2$\pm$0.2 & 89.9 & 99.8 & 99.2 \\ \cline{2-8}
     ACC (unit:\%)\cite{Zellinger2016Central}& A vs. D	& 63.8 & 66.8 & 79.6$\pm$0.6 & - & 73.1 & - \\ \cline{2-8}
     & D vs. A	& 51.1 & 52.8 & 63.8$\pm$0.7 & - & 59.8 & - \\ \cline{2-8}
     & W vs. A	& 49.8 & 51.5 & 63.3$\pm$0.6 & - & 57.4 & - \\ \cline{2-8}
     & Avg      & 70.1 & 72.1 & 79.9	     & - & 76.7 & - \\ \hline
	 & & \textbf{AlexNet} &	\textbf{DLID} & \textbf{DANN} & \textbf{Soft Labels} & \tabincell{l}{Domain\\Confusion} & \tabincell{l}{Confusion\\+Soft} \\ \cline{2-8}
     &A vs. W & 56.5$\pm$0.3 & 51.9 & 53.6$\pm$0.2 & 82.7$\pm$0.7 & 82.8$\pm$0.9 & 82.7$\pm$0.8 \\ \cline{2-8}
     &D vs. W & 92.4$\pm$0.3 & 78.2 & 71.2$\pm$0.0 & 95.9$\pm$0.6 & 95.6$\pm$0.4 & 95.7$\pm$0.5 \\ \cline{2-8}
     Office-31 Dataset &W vs. D & 93.6$\pm$0.2 & 89.9 & 83.5$\pm$0.0 & 98.3$\pm$0.3 & 97.5$\pm$0.2 & 97.6$\pm$0.2 \\ \cline{2-8}
     ACC (unit:\%)\cite{Tzeng2015Simultaneous} &A vs. D & 64.6$\pm$0.4 & -    &-             & 84.9$\pm$1.2 & 85.9$\pm$1.1 & 86.1$\pm$1.2 \\ \cline{2-8}
     &D vs. A & 47.6$\pm$0.1 & -    &-             & 66.0$\pm$0.5 & 66.2$\pm$0.4 & 66.2$\pm$0.3 \\ \cline{2-8}
     &W vs. A & 42.7$\pm$0.1 & -    &-             & 65.2$\pm$0.6 & 64.9$\pm$0.5 & 65.0$\pm$0.5 \\ \cline{2-8}
     &Avg	   & 66.2         & -    &-             & 82.17        & 82.13        & 82.22 \\ \hline
	 MNIST, USPS, & & \textbf{VGG-16} & \textbf{DANN} & \textbf{CoGAN} & \textbf{ADDA} & & \\ \cline{2-8}
     and SVHN&M vs. U & 75.2$\pm$1.6 & 77.1$\pm$1.8 & 91.2$\pm$0.8 & 89.4$\pm$0.2 & & \\ \cline{2-8}
     digits datasets&U vs. M & 57.1$\pm$1.7 & 73.0$\pm$2.0 & 89.1$\pm$0.8 & 90.1$\pm$0.8 & & \\ \cline{2-8}
     ACC (unit:\%)\cite{Tzeng2017Adversarial}&S vs. M & 60.1$\pm$1.1 & 73.9         & -            & 76.0$\pm$1.8 & & \\ \hline

\end{tabular}
\label{tab5}
\end{table*}

Because image classification is a basic task of computer vision applications, most of the algorithms mentioned above were originally proposed to solve such problems. Therefore, we do not discuss this application repeatedly, but we show how much benefit deep DA methods for image classification can bring. Because different papers often use different parameters, experimental protocols and tuning strategies in the preprocessing steps, it is quite difficult to perform a fair comparison among all the methods directly. Thus, similar to the work of Pan \cite{Pan2010A}, we show the comparison results between the proposed deep DA methods and non-adaptation methods using only deep networks. A list of simple experiments taken from some published deep DA papers are presented in Table \ref{tab5}.

In \cite{Long2016Deep}, \cite{Zellinger2016Central}, and \cite{Tzeng2015Simultaneous}, the authors used the Office-31 dataset\footnote{ https://cs.stanford.edu/$\sim$jhoffman/domainadapt/} as one of the evaluation data sets, as shown in Fig. \ref{fig10}(a). The Office dataset is a computer vision classification data set with images from three distinct domains: Amazon (A), DSLR (D), and Webcam (W). The largest domain, Amazon, has 2817 labeled images and its corresponding 31 classes, which consists of objects commonly encountered in office settings. By using this dataset, previous works can show the performance of methods across all six possible DA tasks. \cite{Long2016Deep} showed comparison experiments among the standard AlexNet \cite{Krizhevsky2012ImageNet}, the DANN method \cite{Ganin2015Unsupervised}, and the MMD algorithm and its variations, such as DDC \cite{Tzeng2014Deep}, DAN \cite{Long2015Learning}, JAN \cite{Long2016Deep} and RTN \cite{Long2016Unsupervised}. Zellinger et al. \cite{Zellinger2016Central} evaluated their proposed CMD algorithm in comparison to other discrepancy-based methods (DDC, deep CROAL \cite{Sun2016Deep}, DLID \cite{chopra2013dlid}, AdaBN \cite{Li2016Revisiting}) and the adversarial-based method DANN. \cite{Tzeng2015Simultaneous} proposed an algorithm combining soft label loss and domain confusion loss, and they also compared them with DANN and DLID under a supervised DA setting.

In \cite{Tzeng2017Adversarial}, MNIST\footnote{ http://yann.lecun.com/exdb/mnist/}(M), USPS\footnote{http://statweb.stanford.edu/$\sim$tibs/ElemStatLearn/data.html}(U), and SVHN\footnote{http://ufldl.stanford.edu/housenumbers/} (S) digit datasets (shown in Fig. \ref{fig10}(b)) are used for a cross-domain hand-written digit recognition task, and the experiment showed the comparison results on some adversarial-based methods, such as DANN, CoGAN \cite{Liu2016Coupled} and ADDA \cite{Tzeng2017Adversarial}, where the baseline is VGG-16 \cite{Simonyan2014Very}.

\subsection{Face Recognition}

The performance of face recognition significantly degrades when there are variations in the test images that are not present in the training images. The dataset shift can be caused by poses, resolution, illuminations, expressions, and modality. Kan et al. \cite{Kan2015Bi} proposed a bi-shifting auto-encoder network (BAE) for face recognition across view angle, ethnicity, and imaging sensor. In BAE, source domain samples are shifted to the target domain, and sparse reconstruction is used with several local neighbors from the target domain to ensure its correction, and vice versa. Single sample per person domain adaptation network (SSPP-DAN) in \cite{Hong2017SSPP} generates synthetic images with varying poses to increase the number of samples in the source domain and bridges the gap between the synthetic and source domains by adversarial training with a GRL in real-world face recognition. \cite{Sohn2017Unsupervised} improved the performance of video face recognition by using an adversarial-based approach with large-scale unlabeled videos, labeled still images and synthesized images. Considering that age variations are difficult problems for smile detection and that networks trained on the current benchmarks do not perform well on young children, Xia et al. \cite{Xia_2017_ICCV} applied DAN \cite{Long2015Learning} and JAN \cite{Long2016Deep} (mentioned in Section \ref{Discrepancy-Supervised}) to two baseline deep models, i.e., AlexNet and ResNet, to transfer the knowledge from adults to infants.

\begin{figure}[htbp]
\centering
\includegraphics[width=8cm]{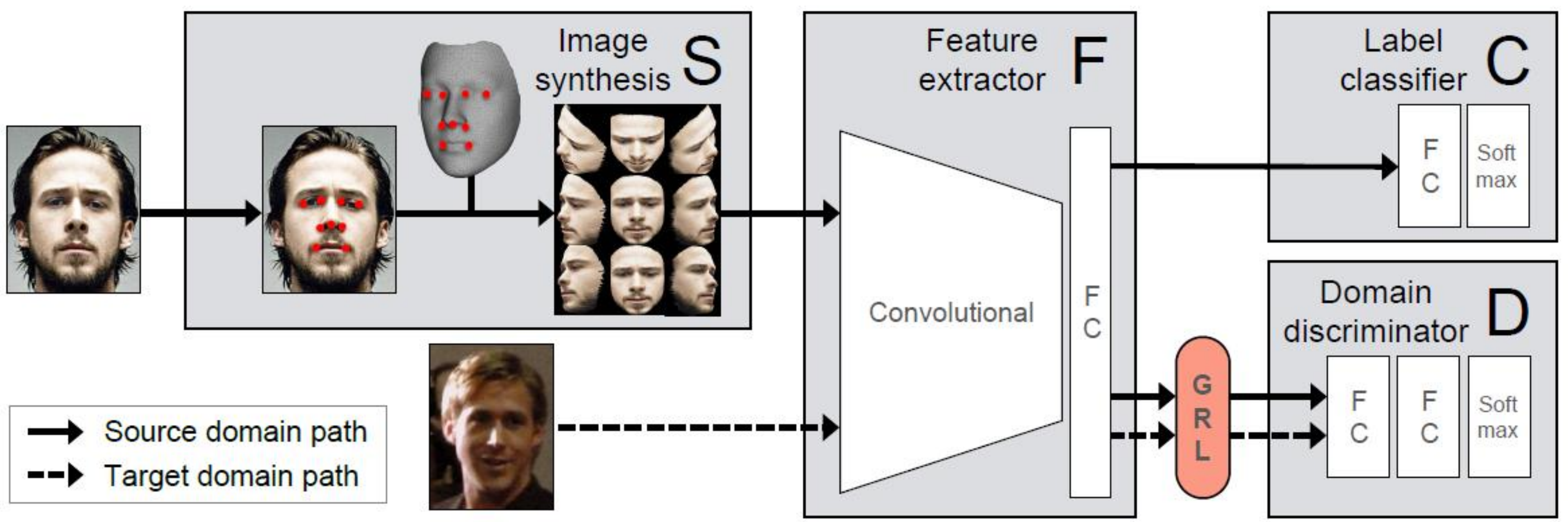}
\caption{The single sample per person domain adaptation network (SSPP-DAN) architecture. \cite{Hong2017SSPP}}
\end{figure}

\subsection{Object Detection}

Recent advances in object detection are driven by region-based convolutional neural networks (R-CNNs \cite{Girshick2013Rich}, fast R-CNNs \cite{Girshick2015Fast} and faster R-CNNs \cite{Ren2015Faster}). They are composed of a window selection mechanism and classifiers that are pre-trained labeled bounding boxes by using the features extracted from CNNs. At test time, the classifier decides whether a region obtained by sliding windows contains the object. Although the R-CNN algorithm is effective, a large amount of bounding box labeled data is required to train each detection category. To solve the problem of lacking labeled data, considering the window selection mechanism as being domain independent, deep DA methods can be used in classifiers to adapt to the target domain.

Because R-CNNs train classifiers on regions just like classification, weak labeled data (such as image-level class labels) are directly useful for the detector. Most works learn the detector with limited bounding box labeled data and massive weak labeled data. The large-scale detection through adaptation (LSDA) \cite{Hoffman2014LSDA} trains a classification layer for the target domain and then uses a pre-trained source model along with output layer adaptation techniques to update the target classification parameters directly. Rochan et al. \cite{Rochan2015Weakly} used word vectors to establish the semantic relatedness between weak labeled source objects and target objects and then transferred the bounding box labeled information from source objects to target objects based on their relatedness. Extending \cite{Hoffman2014LSDA} and \cite{Rochan2015Weakly}, Tang et al. \cite{Tang2016Large} transferred visual (based on the LSDA model) and semantic similarity (based on work vectors) for training an object detector on weak labeled category. \cite{chen2018domain} incorporated both an image-level and an instance-level adaptation component into faster R-CNN and minimized the domain discrepancy based on adversarial training. By using bounding box labeled data in a source domain and weak labeled data in a target domain, \cite{inoue2018cross} progressively fine-tuned the pre-trained model with domain-transfer samples and pseudo-labeling samples.

\subsection{Semantic Segmentation}

Fully convolutional network models (FCNs) for dense prediction have proven to be successful for evaluating semantic segmentation, but their performance will also degrade under domain shifts. Therefore, some work has also explored using weak labels to improve the performance of semantic segmentation. Hong et al. \cite{Hong2015Learning} used a novel encoder-decoder architecture with attention model by transferring weak class labeled knowledge in the source domain, while \cite{Kolesnikov2016Seed,Shimoda2016Distinct} transferred weak object location knowledge.

Much attention has also been paid to deep unsupervised DA in semantic segmentation. Hoffman et al. \cite{Hoffman2016FCNs} first introduced it, in which global domain alignment is performed using FCNs with adversarial-based training, while transferring spatial layout is achieved by leveraging class-aware constrained multiple instance loss. Zhang et al. \cite{Zhang2017Curriculum} enhanced the segmentation performance on real images with the help of virtual ones. It uses the global label distribution loss of the images and local label distribution loss of the landmark superpixels in the target domain to effectively regularize the fine-tuning of the semantic segmentation network. Chen et al. \cite{Chen2017No} proposed a framework for cross-city semantic segmentation. The framework assigns pseudo labels to pixels/grids in the target domain and jointly utilizes global and class-wise alignment by domain adversarial learning to minimize domain shift. In \cite{chen2017road}, a target guided distillation module adapts the style from the real images by imitating the pre-trained source network, and a spatial-aware adaptation module leverages the intrinsic spatial structure to reduce the domain divergence. Rather than operating a simple adversarial objective on the feature space, \cite{Sankaranarayanan2017Learning} used a GAN to address domain shift in which a generator projects the features to the image space and a discriminator operates on this projected image space.

\begin{figure}[htbp]
\centering
\includegraphics[width=8cm]{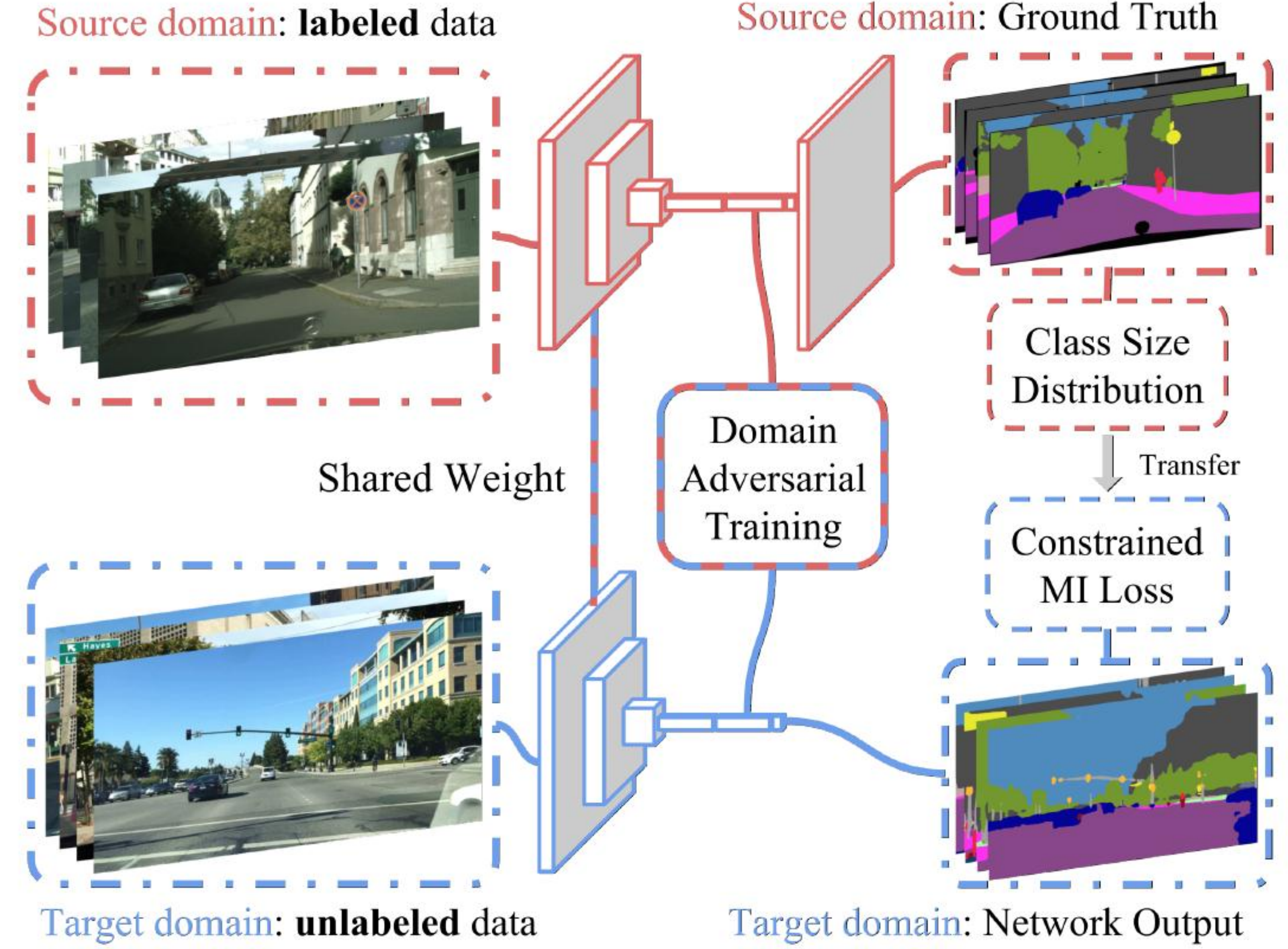}
\caption{The architecture of pixel-level adversarial and constraint-based adaptation. \cite{Hoffman2016FCNs}}
\end{figure}

\subsection{Image-to-Image Translation}

Image-to-image translation has recently achieved great success with deep DA, and it has been applied to various tasks, such as style transferring. Specially, when the feature spaces of source and target images are not same, image-to-image translation should be performed by heterogeneous DA.

More approaches of image-to-image translation use a dataset of paired images and incorporate a DA algorithm into generative networks. Isola et al. \cite{Isola2016Image} proposed the pix2pix framework, which uses a conditional GAN to learn a mapping from source to target images. Tzeng et al. \cite{Tzeng2015Adapting} utilized domain confusion loss and pairwise loss to adapt from simulation to real-world data in a PR2 robot. However, several other methods also address the unpaired setting, such as CoGAN \cite{Liu2016Coupled}, cycle GAN \cite{Zhu2017Unpaired}, dual GAN \cite{Yi2017DualGAN} and disco GAN \cite{Kim2017Learning}.

Matching the statistical distribution by fine-tuning a deep network is another way to achieve image-to-image translation. Gatys et al. \cite{Gatys2016Image} fine-tuned the CNN to achieve DA by the total loss, which is a linear combination between the content and the style loss, such that the target image is rendered in the style of the source image maintaining the content. The content loss minimizes the mean squared difference of the feature representation between the original image and generated image in higher layers, while the style loss minimizes the element-wise mean squared difference between the Gram matrix of them on each layer. \cite{Li2017Demystifying} demonstrated that matching the Gram matrices of feature maps is equivalent to minimizing the MMD. Rather than MMD, \cite{Peng2017Synthetic} proposed a deep generative correlation alignment network (DGCAN) that bridges the domain discrepancy between CAD synthetic and real images by applying the content and CORAL losses to different layers.

\subsection{Person Re-identification}

In the community, person re-identification (re-ID) has become increasingly popular. When given video sequences of a person, person re-ID recognizes whether this person has been in another camera to compensate for the limitations of fixed devices. Recently, deep DA methods have been used in re-ID when models trained on one dataset are directly used on another. Xiao et al. \cite{Xiao2016Learning} proposed the domain-guided dropout algorithm to discard useless neurons for re-identifying persons on multiple datasets simultaneously. Inspired by cycle GAN and Siamese network, the similarity preserving generative adversarial network (SPGAN) \cite{deng2017image} translated the labeled source image to the target domain, preserving self similarity and domain-dissimilarity in an unsupervised manner, and then it trains re-ID models with the translated images using supervised feature learning methods.

\subsection{Image Captioning}

Recently, image captioning, which automatically describes an image with a natural sentence, has been an emerging challenge in computer vision and natural language processing. Due to lacking of paired image-sentence training data, DA leverages different types of data in other source domains to tackle this challenge. Chen et al. \cite{chen2017show} proposed a novel adversarial training procedure (captioner v.s. critics) for cross-domain image captioning using paired source data and unpaired target data. One captioner adapts the sentence style from source to target domain, whereas two critics, namely domain critic and multi-modal critic, aim at distinguishing them. Zhao et al. \cite{zhao2017dual} fine-tuned the pre-trained source model on limited data in the target domain via a dual learning mechanism.

\section{Conclusion}

In a broad sense, deep DA is utilizing deep networks to enhance the performance of DA, such as shallow DA methods with features extracted by deep networks. In a narrow sense, deep DA is based on deep learning architectures designed for DA and optimized by back propagation. In this survey paper, we focus on this narrow definition, and we have reviewed deep DA techniques on visual categorization tasks.

Deep DA is classified as homogeneous DA and heterogeneous DA, and it can be further divided into supervised, semi-supervised and unsupervised settings. The first setting is the simplest but is generally limited due to the need for labeled data; thus, most previous works focused on unsupervised cases. Semi-supervised deep DA is a hybrid method that combines the methods of the supervised and unsupervised settings.

Furthermore, the approaches of deep DA can be classified into one-step DA and multi-step DA considering the distance of the source and target domains. When the distance is small, one-step DA can be used based on training loss. It consists of the discrepancy-based approach, the adversarial-based approach, and the reconstruction-based approach. When the source and target domains are not directly related, multi-step (or transitive) DA can be used. The key of multi-step DA is to select and utilize intermediate domains, thus falling into three categories, including hand-crafted, feature-based and representation-based selection mechanisms.

Although deep DA has achieved success recently, many issues still remain to be addressed. First, most existing algorithms focus on homogeneous deep DA, which assumes that the feature spaces between the source and target domains are the same. However, this assumption may not be true in many applications. We expect to transfer knowledge without this severe limitation and take advantage of existing datasets to help with more tasks. Heterogeneous deep DA may attract increasingly more attention in the future.

In addition, deep DA techniques have been successfully applied in many real-world applications, including image classification, and style translation. We have also found that only a few papers address adaptation beyond classification and recognition, such as object detection, face recognition, semantic segmentation and person re-identification. How to achieve these tasks with no or a very limited amount of data is probably one of the main challenges that should be addressed by deep DA in the next few years.

Finally, since existing deep DA methods aim at aligning marginal distributions, they commonly assume shared label space across the source and target domains. However, in realistic scenario, the images of the source and target domain may be from the different set of categories or only a few categories of interest are shared. Recently, some papers \cite{Cao2017Partial,busto2017open,zhang2018importance} have begun to focus on this issue and we believe it is worthy of more attention.

\section{Acknowledgements}
This work was partially supported by the National Natural Science Foundation of China under Grant Nos. 61573068, 61471048, and 61375031, and Beijing Nova Program under Grant No. Z161100004916088.

{\footnotesize
\bibliographystyle{ieee}
\bibliography{1toN}

\begin{thebibliography}{100}\itemsep=-1pt

\bibitem{arjovsky2017wasserstein}
M.~Arjovsky, S.~Chintala, and L.~Bottou.
\newblock Wasserstein gan.
\newblock {\em arXiv preprint arXiv:1701.07875}, 2017.

\bibitem{Bengio2009Learning}
Y.~Bengio.
\newblock Learning deep architectures for ai.
\newblock {\em Foundations and Trends in Machine Learning}, 2(1):1--127, 2009.

\bibitem{Borgwardt2006Integrating}
K.~M. Borgwardt, A.~Gretton, M.~J. Rasch, H.-P. Kriegel, B.~Sch{\"o}lkopf, and
  A.~J. Smola.
\newblock Integrating structured biological data by kernel maximum mean
  discrepancy.
\newblock {\em Bioinformatics}, 22(14):e49--e57, 2006.

\bibitem{Bousmalis2016Unsupervised}
K.~Bousmalis, N.~Silberman, D.~Dohan, D.~Erhan, and D.~Krishnan.
\newblock Unsupervised pixel-level domain adaptation with generative
  adversarial networks.
\newblock {\em arXiv preprint arXiv:1612.05424}, 2016.

\bibitem{Bousmalis2016Domain}
K.~Bousmalis, G.~Trigeorgis, N.~Silberman, D.~Krishnan, and D.~Erhan.
\newblock Domain separation networks.
\newblock In {\em Advances in Neural Information Processing Systems}, pages
  343--351, 2016.

\bibitem{Bruzzone2010Domain}
L.~Bruzzone and M.~Marconcini.
\newblock Domain adaptation problems: A dasvm classification technique and a
  circular validation strategy.
\newblock {\em IEEE transactions on pattern analysis and machine intelligence},
  32(5):770--787, 2010.

\bibitem{busto2017open}
P.~P. Busto and J.~Gall.
\newblock Open set domain adaptation.
\newblock In {\em The IEEE International Conference on Computer Vision (ICCV)},
  volume~1, page~3, 2017.

\bibitem{Cao2017Partial}
Z.~Cao, M.~Long, J.~Wang, and M.~I. Jordan.
\newblock Partial transfer learning with selective adversarial networks.
\newblock {\em arXiv preprint arXiv:1707.07901}, 2017.

\bibitem{carlucci2017autodial}
F.~M. Carlucci, L.~Porzi, B.~Caputo, E.~Ricci, and S.~R. Bul{\`o}.
\newblock Autodial: Automatic domain alignment layers.
\newblock In {\em International Conference on Computer Vision}, 2017.

\bibitem{Chen2012Marginalized}
M.~Chen, Z.~Xu, K.~Weinberger, and F.~Sha.
\newblock Marginalized denoising autoencoders for domain adaptation.
\newblock {\em arXiv preprint arXiv:1206.4683}, 2012.

\bibitem{chen2017show}
T.-H. Chen, Y.-H. Liao, C.-Y. Chuang, W.-T. Hsu, J.~Fu, and M.~Sun.
\newblock Show, adapt and tell: Adversarial training of cross-domain image
  captioner.
\newblock In {\em The IEEE International Conference on Computer Vision (ICCV)},
  volume~2, 2017.

\bibitem{Chen2016Transfer}
W.-Y. Chen, T.-M.~H. Hsu, Y.-H.~H. Tsai, Y.-C.~F. Wang, and M.-S. Chen.
\newblock Transfer neural trees for heterogeneous domain adaptation.
\newblock In {\em European Conference on Computer Vision}, pages 399--414.
  Springer, 2016.

\bibitem{chen2018domain}
Y.~Chen, W.~Li, C.~Sakaridis, D.~Dai, and L.~Van~Gool.
\newblock Domain adaptive faster r-cnn for object detection in the wild.
\newblock {\em arXiv preprint arXiv:1803.03243}, 2018.

\bibitem{chen2017road}
Y.~Chen, W.~Li, and L.~Van~Gool.
\newblock Road: Reality oriented adaptation for semantic segmentation of urban
  scenes.
\newblock {\em arXiv preprint arXiv:1711.11556}, 2017.

\bibitem{Chen2017No}
Y.-H. Chen, W.-Y. Chen, Y.-T. Chen, B.-C. Tsai, Y.-C.~F. Wang, and M.~Sun.
\newblock No more discrimination: Cross city adaptation of road scene
  segmenters.
\newblock {\em arXiv preprint arXiv:1704.08509}, 2017.

\bibitem{chopra2013dlid}
S.~Chopra, S.~Balakrishnan, and R.~Gopalan.
\newblock Dlid: Deep learning for domain adaptation by interpolating between
  domains.
\newblock In {\em ICML workshop on challenges in representation learning},
  volume~2, 2013.

\bibitem{Chu2016Best}
B.~Chu, V.~Madhavan, O.~Beijbom, J.~Hoffman, and T.~Darrell.
\newblock Best practices for fine-tuning visual classifiers to new domains.
\newblock In {\em Computer Vision--ECCV 2016 Workshops}, pages 435--442.
  Springer, 2016.

\bibitem{Chu2017Selective}
W.-S. Chu, F.~De~la Torre, and J.~F. Cohn.
\newblock Selective transfer machine for personalized facial action unit
  detection.
\newblock In {\em Proceedings of the IEEE Conference on Computer Vision and
  Pattern Recognition}, pages 3515--3522, 2013.

\bibitem{Csurka2017Domain}
G.~Csurka.
\newblock Domain adaptation for visual applications: A comprehensive survey.
\newblock {\em arXiv preprint arXiv:1702.05374}, 2017.

\bibitem{Day2017A}
O.~Day and T.~M. Khoshgoftaar.
\newblock A survey on heterogeneous transfer learning.
\newblock {\em Journal of Big Data}, 4(1):29, 2017.

\bibitem{deng2017image}
W.~Deng, L.~Zheng, G.~Kang, Y.~Yang, Q.~Ye, and J.~Jiao.
\newblock Image-image domain adaptation with preserved self-similarity and
  domain-dissimilarity for person re-identification.
\newblock {\em arXiv preprint arXiv:1711.07027}, 2017.

\bibitem{Donahue2013DeCAF}
J.~Donahue, Y.~Jia, O.~Vinyals, J.~Hoffman, N.~Zhang, E.~Tzeng, and T.~Darrell.
\newblock Decaf: A deep convolutional activation feature for generic visual
  recognition.
\newblock In {\em International conference on machine learning}, pages
  647--655, 2014.

\bibitem{duan2012learning}
L.~Duan, D.~Xu, and I.~Tsang.
\newblock Learning with augmented features for heterogeneous domain adaptation.
\newblock {\em arXiv preprint arXiv:1206.4660}, 2012.

\bibitem{Eigen2014Depth}
D.~Eigen, C.~Puhrsch, and R.~Fergus.
\newblock Depth map prediction from a single image using a multi-scale deep
  network.
\newblock In {\em Advances in neural information processing systems}, pages
  2366--2374, 2014.

\bibitem{Ganin2015Unsupervised}
Y.~Ganin and V.~Lempitsky.
\newblock Unsupervised domain adaptation by backpropagation.
\newblock In {\em International Conference on Machine Learning}, pages
  1180--1189, 2015.

\bibitem{Ganin2017Domain}
Y.~Ganin, E.~Ustinova, H.~Ajakan, P.~Germain, H.~Larochelle, F.~Laviolette,
  M.~Marchand, and V.~Lempitsky.
\newblock Domain-adversarial training of neural networks.
\newblock {\em Journal of Machine Learning Research}, 17(59):1--35, 2016.

\bibitem{Gatys2016Image}
L.~A. Gatys, A.~S. Ecker, and M.~Bethge.
\newblock Image style transfer using convolutional neural networks.
\newblock In {\em Proceedings of the IEEE Conference on Computer Vision and
  Pattern Recognition}, pages 2414--2423, 2016.

\bibitem{Ge2017Borrowing}
W.~Ge and Y.~Yu.
\newblock Borrowing treasures from the wealthy: Deep transfer learning through
  selective joint fine-tuning.
\newblock {\em arXiv preprint arXiv:1702.08690}, 2017.

\bibitem{Gebru2017Fine}
T.~Gebru, J.~Hoffman, and L.~Fei-Fei.
\newblock Fine-grained recognition in the wild: A multi-task domain adaptation
  approach.
\newblock {\em arXiv preprint arXiv:1709.02476}, 2017.

\bibitem{gheisari2015unsupervised}
M.~Gheisari and M.~S. Baghshah.
\newblock Unsupervised domain adaptation via representation learning and
  adaptive classifier learning.
\newblock {\em Neurocomputing}, 165:300--311, 2015.

\bibitem{Ghifary2015Domain}
M.~Ghifary, W.~Bastiaan~Kleijn, M.~Zhang, and D.~Balduzzi.
\newblock Domain generalization for object recognition with multi-task
  autoencoders.
\newblock In {\em Proceedings of the IEEE international conference on computer
  vision}, pages 2551--2559, 2015.

\bibitem{Ghifary2014Domain}
M.~Ghifary, W.~B. Kleijn, and M.~Zhang.
\newblock Domain adaptive neural networks for object recognition.
\newblock In {\em Pacific Rim International Conference on Artificial
  Intelligence}, pages 898--904. Springer, 2014.

\bibitem{Ghifary2016Deep}
M.~Ghifary, W.~B. Kleijn, M.~Zhang, D.~Balduzzi, and W.~Li.
\newblock Deep reconstruction-classification networks for unsupervised domain
  adaptation.
\newblock In {\em European Conference on Computer Vision}, pages 597--613.
  Springer, 2016.

\bibitem{Girshick2015Fast}
R.~Girshick.
\newblock Fast r-cnn.
\newblock In {\em Proceedings of the IEEE international conference on computer
  vision}, pages 1440--1448, 2015.

\bibitem{Girshick2013Rich}
R.~Girshick, J.~Donahue, T.~Darrell, and J.~Malik.
\newblock Rich feature hierarchies for accurate object detection and semantic
  segmentation.
\newblock In {\em Proceedings of the IEEE conference on computer vision and
  pattern recognition}, pages 580--587, 2014.

\bibitem{Glorot2011Domain}
X.~Glorot, A.~Bordes, and Y.~Bengio.
\newblock Domain adaptation for large-scale sentiment classification: A deep
  learning approach.
\newblock In {\em Proceedings of the 28th international conference on machine
  learning (ICML-11)}, pages 513--520, 2011.

\bibitem{Gong2013Connecting}
B.~Gong, K.~Grauman, and F.~Sha.
\newblock Connecting the dots with landmarks: Discriminatively learning
  domain-invariant features for unsupervised domain adaptation.
\newblock In {\em International Conference on Machine Learning}, pages
  222--230, 2013.

\bibitem{Grauman2012Geodesic}
B.~Gong, Y.~Shi, F.~Sha, and K.~Grauman.
\newblock Geodesic flow kernel for unsupervised domain adaptation.
\newblock In {\em Computer Vision and Pattern Recognition (CVPR), 2012 IEEE
  Conference on}, pages 2066--2073. IEEE, 2012.

\bibitem{Goodfellow2014Generative}
I.~Goodfellow, J.~Pouget-Abadie, M.~Mirza, B.~Xu, D.~Warde-Farley, S.~Ozair,
  A.~Courville, and Y.~Bengio.
\newblock Generative adversarial nets.
\newblock In {\em Advances in neural information processing systems}, pages
  2672--2680, 2014.

\bibitem{Gopalan2011Domain}
R.~Gopalan, R.~Li, and R.~Chellappa.
\newblock Domain adaptation for object recognition: An unsupervised approach.
\newblock In {\em Computer Vision (ICCV), 2011 IEEE International Conference
  on}, pages 999--1006. IEEE, 2011.

\bibitem{Gupta2016Cross}
S.~Gupta, J.~Hoffman, and J.~Malik.
\newblock Cross modal distillation for supervision transfer.
\newblock In {\em Proceedings of the IEEE Conference on Computer Vision and
  Pattern Recognition}, pages 2827--2836, 2016.

\bibitem{haeusser2017associative}
P.~Haeusser, T.~Frerix, A.~Mordvintsev, and D.~Cremers.
\newblock Associative domain adaptation.
\newblock In {\em International Conference on Computer Vision (ICCV)},
  volume~2, page~6, 2017.

\bibitem{Xia2016Dual}
D.~He, Y.~Xia, T.~Qin, L.~Wang, N.~Yu, T.~Liu, and W.-Y. Ma.
\newblock Dual learning for machine translation.
\newblock In {\em Advances in Neural Information Processing Systems}, pages
  820--828, 2016.

\bibitem{He2016Deep}
K.~He, X.~Zhang, S.~Ren, and J.~Sun.
\newblock Deep residual learning for image recognition.
\newblock In {\em Proceedings of the IEEE conference on computer vision and
  pattern recognition}, pages 770--778, 2016.

\bibitem{Hinton2015Distilling}
G.~Hinton, O.~Vinyals, and J.~Dean.
\newblock Distilling the knowledge in a neural network.
\newblock {\em arXiv preprint arXiv:1503.02531}, 2015.

\bibitem{Hinton2006A}
G.~E. Hinton, S.~Osindero, and Y.-W. Teh.
\newblock A fast learning algorithm for deep belief nets.
\newblock {\em Neural computation}, 18(7):1527--1554, 2006.

\bibitem{Hoffman2014LSDA}
J.~Hoffman, S.~Guadarrama, E.~S. Tzeng, R.~Hu, J.~Donahue, R.~Girshick,
  T.~Darrell, and K.~Saenko.
\newblock Lsda: Large scale detection through adaptation.
\newblock In {\em Advances in Neural Information Processing Systems}, pages
  3536--3544, 2014.

\bibitem{Hoffman2016Cross}
J.~Hoffman, S.~Gupta, J.~Leong, S.~Guadarrama, and T.~Darrell.
\newblock Cross-modal adaptation for rgb-d detection.
\newblock In {\em Robotics and Automation (ICRA), 2016 IEEE International
  Conference on}, pages 5032--5039. IEEE, 2016.

\bibitem{Hoffman2013One}
J.~Hoffman, E.~Tzeng, J.~Donahue, Y.~Jia, K.~Saenko, and T.~Darrell.
\newblock One-shot adaptation of supervised deep convolutional models.
\newblock {\em arXiv preprint arXiv:1312.6204}, 2013.

\bibitem{Hoffman2016FCNs}
J.~Hoffman, D.~Wang, F.~Yu, and T.~Darrell.
\newblock Fcns in the wild: Pixel-level adversarial and constraint-based
  adaptation.
\newblock {\em arXiv preprint arXiv:1612.02649}, 2016.

\bibitem{Hong2017SSPP}
S.~Hong, W.~Im, J.~Ryu, and H.~S. Yang.
\newblock Sspp-dan: Deep domain adaptation network for face recognition with
  single sample per person.
\newblock {\em arXiv preprint arXiv:1702.04069}, 2017.

\bibitem{Hong2015Learning}
S.~Hong, J.~Oh, H.~Lee, and B.~Han.
\newblock Learning transferrable knowledge for semantic segmentation with deep
  convolutional neural network.
\newblock In {\em Proceedings of the IEEE Conference on Computer Vision and
  Pattern Recognition}, pages 3204--3212, 2016.

\bibitem{Hu2015Deep}
J.~Hu, J.~Lu, and Y.-P. Tan.
\newblock Deep transfer metric learning.
\newblock In {\em Proceedings of the IEEE Conference on Computer Vision and
  Pattern Recognition}, pages 325--333, 2015.

\bibitem{Huang2017Arbitrary}
X.~Huang and S.~Belongie.
\newblock Arbitrary style transfer in real-time with adaptive instance
  normalization.
\newblock {\em arXiv preprint arXiv:1703.06868}, 2017.

\bibitem{inoue2018cross}
N.~Inoue, R.~Furuta, T.~Yamasaki, and K.~Aizawa.
\newblock Cross-domain weakly-supervised object detection through progressive
  domain adaptation.
\newblock {\em arXiv preprint arXiv:1803.11365}, 2018.

\bibitem{Ioffe2015Batch}
S.~Ioffe and C.~Szegedy.
\newblock Batch normalization: Accelerating deep network training by reducing
  internal covariate shift.
\newblock In {\em International Conference on Machine Learning}, pages
  448--456, 2015.

\bibitem{Isola2016Image}
P.~Isola, J.-Y. Zhu, T.~Zhou, and A.~A. Efros.
\newblock Image-to-image translation with conditional adversarial networks.
\newblock {\em arXiv preprint arXiv:1611.07004}, 2016.

\bibitem{Kan2015Bi}
M.~Kan, S.~Shan, and X.~Chen.
\newblock Bi-shifting auto-encoder for unsupervised domain adaptation.
\newblock In {\em Proceedings of the IEEE International Conference on Computer
  Vision}, pages 3846--3854, 2015.

\bibitem{Kim2017Learning}
T.~Kim, M.~Cha, H.~Kim, J.~Lee, and J.~Kim.
\newblock Learning to discover cross-domain relations with generative
  adversarial networks.
\newblock {\em arXiv preprint arXiv:1703.05192}, 2017.

\bibitem{Kolesnikov2016Seed}
A.~Kolesnikov and C.~H. Lampert.
\newblock Seed, expand and constrain: Three principles for weakly-supervised
  image segmentation.
\newblock In {\em European Conference on Computer Vision}, pages 695--711.
  Springer, 2016.

\bibitem{Kontschieder2016Deep}
P.~Kontschieder, M.~Fiterau, A.~Criminisi, and S.~Rota~Bulo.
\newblock Deep neural decision forests.
\newblock In {\em Proceedings of the IEEE International Conference on Computer
  Vision}, pages 1467--1475, 2015.

\bibitem{Krizhevsky2012ImageNet}
A.~Krizhevsky, I.~Sutskever, and G.~E. Hinton.
\newblock Imagenet classification with deep convolutional neural networks.
\newblock In {\em Advances in neural information processing systems}, pages
  1097--1105, 2012.

\bibitem{kulis2011you}
B.~Kulis, K.~Saenko, and T.~Darrell.
\newblock What you saw is not what you get: Domain adaptation using asymmetric
  kernel transforms.
\newblock In {\em Computer Vision and Pattern Recognition (CVPR), 2011 IEEE
  Conference on}, pages 1785--1792. IEEE, 2011.

\bibitem{Lampert2009Learning}
C.~H. Lampert, H.~Nickisch, and S.~Harmeling.
\newblock Learning to detect unseen object classes by between-class attribute
  transfer.
\newblock In {\em Computer Vision and Pattern Recognition, 2009. CVPR 2009.
  IEEE Conference on}, pages 951--958. IEEE, 2009.

\bibitem{Li2016Precomputed}
C.~Li and M.~Wand.
\newblock Precomputed real-time texture synthesis with markovian generative
  adversarial networks.
\newblock In {\em European Conference on Computer Vision}, pages 702--716.
  Springer, 2016.

\bibitem{li2017deeper}
D.~Li, Y.~Yang, Y.-Z. Song, and T.~M. Hospedales.
\newblock Deeper, broader and artier domain generalization.
\newblock In {\em Computer Vision (ICCV), 2017 IEEE International Conference
  on}, pages 5543--5551. IEEE, 2017.

\bibitem{Li2015Generative}
Y.~Li, K.~Swersky, and R.~Zemel.
\newblock Generative moment matching networks.
\newblock In {\em Proceedings of the 32nd International Conference on Machine
  Learning (ICML-15)}, pages 1718--1727, 2015.

\bibitem{Li2017Demystifying}
Y.~Li, N.~Wang, J.~Liu, and X.~Hou.
\newblock Demystifying neural style transfer.
\newblock {\em arXiv preprint arXiv:1701.01036}, 2017.

\bibitem{Li2016Revisiting}
Y.~Li, N.~Wang, J.~Shi, J.~Liu, and X.~Hou.
\newblock Revisiting batch normalization for practical domain adaptation.
\newblock {\em arXiv preprint arXiv:1603.04779}, 2016.

\bibitem{Liu2016Coupled}
M.-Y. Liu and O.~Tuzel.
\newblock Coupled generative adversarial networks.
\newblock In {\em Advances in neural information processing systems}, pages
  469--477, 2016.

\bibitem{liu2017survey}
W.~Liu, Z.~Wang, X.~Liu, N.~Zeng, Y.~Liu, and F.~E. Alsaadi.
\newblock A survey of deep neural network architectures and their applications.
\newblock {\em Neurocomputing}, 234:11--26, 2017.

\bibitem{liu2016transferring}
X.~Liu, L.~Song, X.~Wu, and T.~Tan.
\newblock Transferring deep representation for nir-vis heterogeneous face
  recognition.
\newblock In {\em Biometrics (ICB), 2016 International Conference on}, pages
  1--8. IEEE, 2016.

\bibitem{Long2015Learning}
M.~Long, Y.~Cao, J.~Wang, and M.~Jordan.
\newblock Learning transferable features with deep adaptation networks.
\newblock In {\em International Conference on Machine Learning}, pages 97--105,
  2015.

\bibitem{Long2016Deep}
M.~Long, J.~Wang, and M.~I. Jordan.
\newblock Deep transfer learning with joint adaptation networks.
\newblock {\em arXiv preprint arXiv:1605.06636}, 2016.

\bibitem{Long2016Unsupervised}
M.~Long, H.~Zhu, J.~Wang, and M.~I. Jordan.
\newblock Unsupervised domain adaptation with residual transfer networks.
\newblock In {\em Advances in Neural Information Processing Systems}, pages
  136--144, 2016.

\bibitem{lu2017unsupervised}
H.~Lu, L.~Zhang, Z.~Cao, W.~Wei, K.~Xian, C.~Shen, and A.~van~den Hengel.
\newblock When unsupervised domain adaptation meets tensor representations.
\newblock In {\em The IEEE International Conference on Computer Vision (ICCV)},
  volume~2, 2017.

\bibitem{mittal2015composite}
P.~Mittal, M.~Vatsa, and R.~Singh.
\newblock Composite sketch recognition via deep network-a transfer learning
  approach.
\newblock In {\em Biometrics (ICB), 2015 International Conference on}, pages
  251--256. IEEE, 2015.

\bibitem{motiian2017few}
S.~Motiian, Q.~Jones, S.~Iranmanesh, and G.~Doretto.
\newblock Few-shot adversarial domain adaptation.
\newblock In {\em Advances in Neural Information Processing Systems}, pages
  6673--6683, 2017.

\bibitem{motiian2017unified}
S.~Motiian, M.~Piccirilli, D.~A. Adjeroh, and G.~Doretto.
\newblock Unified deep supervised domain adaptation and generalization.
\newblock In {\em The IEEE International Conference on Computer Vision (ICCV)},
  volume~2, 2017.

\bibitem{Nguyen2015DASH}
H.~V. Nguyen, H.~T. Ho, V.~M. Patel, and R.~Chellappa.
\newblock Dash-n: Joint hierarchical domain adaptation and feature learning.
\newblock {\em IEEE Transactions on Image Processing}, 24(12):5479--5491, 2015.

\bibitem{pachori2017hashing}
S.~Pachori, A.~Deshpande, and S.~Raman.
\newblock Hashing in the zero shot framework with domain adaptation.
\newblock {\em Neurocomputing}, 2017.

\bibitem{Pan2011Domain}
S.~J. Pan, I.~W. Tsang, J.~T. Kwok, and Q.~Yang.
\newblock Domain adaptation via transfer component analysis.
\newblock {\em IEEE Transactions on Neural Networks}, 22(2):199--210, 2011.

\bibitem{Pan2010A}
S.~J. Pan and Q.~Yang.
\newblock A survey on transfer learning.
\newblock {\em IEEE Transactions on knowledge and data engineering},
  22(10):1345--1359, 2010.

\bibitem{Patel2015Visual}
V.~M. Patel, R.~Gopalan, R.~Li, and R.~Chellappa.
\newblock Visual domain adaptation: A survey of recent advances.
\newblock {\em IEEE signal processing magazine}, 32(3):53--69, 2015.

\bibitem{Peng2017Zero}
K.-C. Peng, Z.~Wu, and J.~Ernst.
\newblock Zero-shot deep domain adaptation.
\newblock {\em arXiv preprint arXiv:1707.01922}, 2017.

\bibitem{Peng2016Fine}
X.~Peng, J.~Hoffman, X.~Y. Stella, and K.~Saenko.
\newblock Fine-to-coarse knowledge transfer for low-res image classification.
\newblock In {\em Image Processing (ICIP), 2016 IEEE International Conference
  on}, pages 3683--3687. IEEE, 2016.

\bibitem{Peng2017Synthetic}
X.~Peng and K.~Saenko.
\newblock Synthetic to real adaptation with deep generative correlation
  alignment networks.
\newblock {\em arXiv preprint arXiv:1701.05524}, 2017.

\bibitem{Raj2015Subspace}
A.~Raj, V.~P. Namboodiri, and T.~Tuytelaars.
\newblock Subspace alignment based domain adaptation for rcnn detector.
\newblock {\em arXiv preprint arXiv:1507.05578}, 2015.

\bibitem{Rebuffi2017Learning}
S.-A. Rebuffi, H.~Bilen, and A.~Vedaldi.
\newblock Learning multiple visual domains with residual adapters.
\newblock {\em arXiv preprint arXiv:1705.08045}, 2017.

\bibitem{reed2016generative}
S.~Reed, Z.~Akata, X.~Yan, L.~Logeswaran, B.~Schiele, and H.~Lee.
\newblock Generative adversarial text to image synthesis.
\newblock {\em arXiv preprint arXiv:1605.05396}, 2016.

\bibitem{Ren2015Faster}
S.~Ren, K.~He, R.~Girshick, and J.~Sun.
\newblock Faster r-cnn: Towards real-time object detection with region proposal
  networks.
\newblock In {\em Advances in neural information processing systems}, pages
  91--99, 2015.

\bibitem{Rochan2015Weakly}
M.~Rochan and Y.~Wang.
\newblock Weakly supervised localization of novel objects using appearance
  transfer.
\newblock In {\em Proceedings of the IEEE Conference on Computer Vision and
  Pattern Recognition}, pages 4315--4324, 2015.

\bibitem{Ronneberger2015U}
O.~Ronneberger, P.~Fischer, and T.~Brox.
\newblock U-net: Convolutional networks for biomedical image segmentation.
\newblock In {\em International Conference on Medical Image Computing and
  Computer-Assisted Intervention}, pages 234--241. Springer, 2015.

\bibitem{Kontschieder2014Neural}
S.~Rota~Bulo and P.~Kontschieder.
\newblock Neural decision forests for semantic image labelling.
\newblock In {\em Proceedings of the IEEE Conference on Computer Vision and
  Pattern Recognition}, pages 81--88, 2014.

\bibitem{Rozantsev2016Beyond}
A.~Rozantsev, M.~Salzmann, and P.~Fua.
\newblock Beyond sharing weights for deep domain adaptation.
\newblock {\em arXiv preprint arXiv:1603.06432}, 2016.

\bibitem{Rusu2016Progressive}
A.~A. Rusu, N.~C. Rabinowitz, G.~Desjardins, H.~Soyer, J.~Kirkpatrick,
  K.~Kavukcuoglu, R.~Pascanu, and R.~Hadsell.
\newblock Progressive neural networks.
\newblock {\em arXiv preprint arXiv:1606.04671}, 2016.

\bibitem{saenko2010adapting}
K.~Saenko, B.~Kulis, M.~Fritz, and T.~Darrell.
\newblock Adapting visual category models to new domains.
\newblock In {\em European conference on computer vision}, pages 213--226.
  Springer, 2010.

\bibitem{saito2017asymmetric}
K.~Saito, Y.~Ushiku, and T.~Harada.
\newblock Asymmetric tri-training for unsupervised domain adaptation.
\newblock {\em arXiv preprint arXiv:1702.08400}, 2017.

\bibitem{saito2017maximum}
K.~Saito, K.~Watanabe, Y.~Ushiku, and T.~Harada.
\newblock Maximum classifier discrepancy for unsupervised domain adaptation.
\newblock {\em arXiv preprint arXiv:1712.02560}, 2017.

\bibitem{Sankaranarayanan2017Learning}
S.~Sankaranarayanan, Y.~Balaji, A.~Jain, S.~N. Lim, and R.~Chellappa.
\newblock Learning from synthetic data: Addressing domain shift for semantic
  segmentation.
\newblock 2017.

\bibitem{Shao2015Transfer}
L.~Shao, F.~Zhu, and X.~Li.
\newblock Transfer learning for visual categorization: A survey.
\newblock {\em IEEE transactions on neural networks and learning systems},
  26(5):1019--1034, 2015.

\bibitem{Shen2017Wasserstein}
J.~Shen, Y.~Qu, W.~Zhang, and Y.~Yu.
\newblock Wasserstein distance guided representation learning for domain
  adaptation.
\newblock 2017.

\bibitem{Shimoda2016Distinct}
W.~Shimoda and K.~Yanai.
\newblock Distinct class-specific saliency maps for weakly supervised semantic
  segmentation.
\newblock In {\em European Conference on Computer Vision}, pages 218--234.
  Springer, 2016.

\bibitem{Shrivastava2016Learning}
A.~Shrivastava, T.~Pfister, O.~Tuzel, J.~Susskind, W.~Wang, and R.~Webb.
\newblock Learning from simulated and unsupervised images through adversarial
  training.
\newblock {\em arXiv preprint arXiv:1612.07828}, 2016.

\bibitem{Shu2015Weakly}
X.~Shu, G.-J. Qi, J.~Tang, and J.~Wang.
\newblock Weakly-shared deep transfer networks for heterogeneous-domain
  knowledge propagation.
\newblock In {\em Proceedings of the 23rd ACM international conference on
  Multimedia}, pages 35--44. ACM, 2015.

\bibitem{Simonyan2014Very}
K.~Simonyan and A.~Zisserman.
\newblock Very deep convolutional networks for large-scale image recognition.
\newblock {\em arXiv preprint arXiv:1409.1556}, 2014.

\bibitem{Sohn2017Unsupervised}
K.~Sohn, S.~Liu, G.~Zhong, X.~Yu, M.-H. Yang, and M.~Chandraker.
\newblock Unsupervised domain adaptation for face recognition in unlabeled
  videos.
\newblock {\em arXiv preprint arXiv:1708.02191}, 2017.

\bibitem{Sun2016Return}
B.~Sun, J.~Feng, and K.~Saenko.
\newblock Return of frustratingly easy domain adaptation.
\newblock In {\em AAAI}, volume~6, page~8, 2016.

\bibitem{Sun2016Deep}
B.~Sun and K.~Saenko.
\newblock Deep coral: Correlation alignment for deep domain adaptation.
\newblock In {\em Computer Vision--ECCV 2016 Workshops}, pages 443--450.
  Springer, 2016.

\bibitem{Szegedy2015Going}
C.~Szegedy, W.~Liu, Y.~Jia, P.~Sermanet, S.~Reed, D.~Anguelov, D.~Erhan,
  V.~Vanhoucke, and A.~Rabinovich.
\newblock Going deeper with convolutions.
\newblock In {\em Proceedings of the IEEE conference on computer vision and
  pattern recognition}, pages 1--9, 2015.

\bibitem{Taigman2016Unsupervised}
Y.~Taigman, A.~Polyak, and L.~Wolf.
\newblock Unsupervised cross-domain image generation.
\newblock {\em arXiv preprint arXiv:1611.02200}, 2016.

\bibitem{Taigman2014DeepFace}
Y.~Taigman, M.~Yang, M.~Ranzato, and L.~Wolf.
\newblock Deepface: Closing the gap to human-level performance in face
  verification.
\newblock In {\em Proceedings of the IEEE conference on computer vision and
  pattern recognition}, pages 1701--1708, 2014.

\bibitem{Tan2015Transitive}
B.~Tan, Y.~Song, E.~Zhong, and Q.~Yang.
\newblock Transitive transfer learning.
\newblock In {\em Proceedings of the 21th ACM SIGKDD International Conference
  on Knowledge Discovery and Data Mining}, pages 1155--1164. ACM, 2015.

\bibitem{tan2017distant}
B.~Tan, Y.~Zhang, S.~J. Pan, and Q.~Yang.
\newblock Distant domain transfer learning.
\newblock In {\em AAAI}, pages 2604--2610, 2017.

\bibitem{Tang2016Large}
Y.~Tang, J.~Wang, B.~Gao, E.~Dellandr{\'e}a, R.~Gaizauskas, and L.~Chen.
\newblock Large scale semi-supervised object detection using visual and
  semantic knowledge transfer.
\newblock In {\em Proceedings of the IEEE Conference on Computer Vision and
  Pattern Recognition}, pages 2119--2128, 2016.

\bibitem{tsai2017adversarial}
J.-C. Tsai and J.-T. Chien.
\newblock Adversarial domain separation and adaptation.
\newblock In {\em Machine Learning for Signal Processing (MLSP), 2017 IEEE 27th
  International Workshop on}, pages 1--6. IEEE, 2017.

\bibitem{Tzeng2015Adapting}
E.~Tzeng, C.~Devin, J.~Hoffman, C.~Finn, P.~Abbeel, S.~Levine, K.~Saenko, and
  T.~Darrell.
\newblock Adapting deep visuomotor representations with weak pairwise
  constraints.
\newblock {\em CoRR, vol. abs/1511.07111}, 2015.

\bibitem{Tzeng2015Simultaneous}
E.~Tzeng, J.~Hoffman, T.~Darrell, and K.~Saenko.
\newblock Simultaneous deep transfer across domains and tasks.
\newblock In {\em Proceedings of the IEEE International Conference on Computer
  Vision}, pages 4068--4076, 2015.

\bibitem{Tzeng2017Adversarial}
E.~Tzeng, J.~Hoffman, K.~Saenko, and T.~Darrell.
\newblock Adversarial discriminative domain adaptation.
\newblock {\em arXiv preprint arXiv:1702.05464}, 2017.

\bibitem{Tzeng2014Deep}
E.~Tzeng, J.~Hoffman, N.~Zhang, K.~Saenko, and T.~Darrell.
\newblock Deep domain confusion: Maximizing for domain invariance.
\newblock {\em arXiv preprint arXiv:1412.3474}, 2014.

\bibitem{Ulyanov2017Improved}
D.~Ulyanov, A.~Vedaldi, and V.~Lempitsky.
\newblock Improved texture networks: Maximizing quality and diversity in
  feed-forward stylization and texture synthesis.
\newblock {\em arXiv preprint arXiv:1701.02096}, 2017.

\bibitem{Vincent2010Stacked}
P.~Vincent, H.~Larochelle, I.~Lajoie, Y.~Bengio, and P.-A. Manzagol.
\newblock Stacked denoising autoencoders: Learning useful representations in a
  deep network with a local denoising criterion.
\newblock {\em Journal of Machine Learning Research}, 11(Dec):3371--3408, 2010.

\bibitem{volpi2017adversarial}
R.~Volpi, P.~Morerio, S.~Savarese, and V.~Murino.
\newblock Adversarial feature augmentation for unsupervised domain adaptation.
\newblock {\em arXiv preprint arXiv:1711.08561}, 2017.

\bibitem{wang2011heterogeneous}
C.~Wang and S.~Mahadevan.
\newblock Heterogeneous domain adaptation using manifold alignment.
\newblock In {\em IJCAI proceedings-international joint conference on
  artificial intelligence}, volume~22, page 1541, 2011.

\bibitem{wang2017high}
L.~Wang, V.~A. Sindagi, and V.~M. Patel.
\newblock High-quality facial photo-sketch synthesis using multi-adversarial
  networks.
\newblock {\em arXiv preprint arXiv:1710.10182}, 2017.

\bibitem{wang2016deep}
X.~Wang, X.~Duan, and X.~Bai.
\newblock Deep sketch feature for cross-domain image retrieval.
\newblock {\em Neurocomputing}, 207:387--397, 2016.

\bibitem{Xia_2017_ICCV}
Y.~Xia, D.~Huang, and Y.~Wang.
\newblock Detecting smiles of young children via deep transfer learning.
\newblock In {\em Proceedings of the IEEE Conference on Computer Vision and
  Pattern Recognition}, pages 1673--1681, 2017.

\bibitem{Xiao2016Learning}
T.~Xiao, H.~Li, W.~Ouyang, and X.~Wang.
\newblock Learning deep feature representations with domain guided dropout for
  person re-identification.
\newblock In {\em Proceedings of the IEEE Conference on Computer Vision and
  Pattern Recognition}, pages 1249--1258, 2016.

\bibitem{Xie2015Transfer}
M.~Xie, N.~Jean, M.~Burke, D.~Lobell, and S.~Ermon.
\newblock Transfer learning from deep features for remote sensing and poverty
  mapping.
\newblock 2015.

\bibitem{Yan2017Mind}
H.~Yan, Y.~Ding, P.~Li, Q.~Wang, Y.~Xu, and W.~Zuo.
\newblock Mind the class weight bias: Weighted maximum mean discrepancy for
  unsupervised domain adaptation.
\newblock {\em arXiv preprint arXiv:1705.00609}, 2017.

\bibitem{Yi2017DualGAN}
Z.~Yi, H.~Zhang, P.~T. Gong, et~al.
\newblock Dualgan: Unsupervised dual learning for image-to-image translation.
\newblock {\em arXiv preprint arXiv:1704.02510}, 2017.

\bibitem{Yoo2016Pixel}
D.~Yoo, N.~Kim, S.~Park, A.~S. Paek, and I.~S. Kweon.
\newblock Pixel-level domain transfer.
\newblock In {\em European Conference on Computer Vision}, pages 517--532.
  Springer, 2016.

\bibitem{Yosinski2014How}
J.~Yosinski, J.~Clune, Y.~Bengio, and H.~Lipson.
\newblock How transferable are features in deep neural networks?
\newblock In {\em Advances in neural information processing systems}, pages
  3320--3328, 2014.

\bibitem{Zellinger2016Central}
W.~Zellinger, T.~Grubinger, E.~Lughofer, T.~Natschl{\"a}ger, and
  S.~Saminger-Platz.
\newblock Central moment discrepancy (cmd) for domain-invariant representation
  learning.
\newblock {\em arXiv preprint arXiv:1702.08811}, 2017.

\bibitem{zhang2017stackgan}
H.~Zhang, T.~Xu, H.~Li, S.~Zhang, X.~Huang, X.~Wang, and D.~Metaxas.
\newblock Stackgan: Text to photo-realistic image synthesis with stacked
  generative adversarial networks.
\newblock In {\em IEEE Int. Conf. Comput. Vision (ICCV)}, pages 5907--5915,
  2017.

\bibitem{zhang2018importance}
J.~Zhang, Z.~Ding, W.~Li, and P.~Ogunbona.
\newblock Importance weighted adversarial nets for partial domain adaptation.
\newblock {\em arXiv preprint arXiv:1803.09210}, 2018.

\bibitem{Zhang2017Transfer}
J.~Zhang, W.~Li, and P.~Ogunbona.
\newblock Transfer learning for cross-dataset recognition: A survey.
\newblock 2017.

\bibitem{zhang2017deep}
L.~Zhang, Z.~He, and Y.~Liu.
\newblock Deep object recognition across domains based on adaptive extreme
  learning machine.
\newblock {\em Neurocomputing}, 239:194--203, 2017.

\bibitem{Zhang2015Deep}
X.~Zhang, F.~X. Yu, S.-F. Chang, and S.~Wang.
\newblock Deep transfer network: Unsupervised domain adaptation.
\newblock {\em arXiv preprint arXiv:1503.00591}, 2015.

\bibitem{Zhang2017Curriculum}
Y.~Zhang, P.~David, and B.~Gong.
\newblock Curriculum domain adaptation for semantic segmentation of urban
  scenes.
\newblock In {\em The IEEE International Conference on Computer Vision (ICCV)},
  volume~2, page~6, 2017.

\bibitem{zhao2017dual}
W.~Zhao, W.~Xu, M.~Yang, J.~Ye, Z.~Zhao, Y.~Feng, and Y.~Qiao.
\newblock Dual learning for cross-domain image captioning.
\newblock In {\em Proceedings of the 2017 ACM on Conference on Information and
  Knowledge Management}, pages 29--38. ACM, 2017.

\bibitem{zhou2014heterogeneous}
J.~T. Zhou, I.~W. Tsang, S.~J. Pan, and M.~Tan.
\newblock Heterogeneous domain adaptation for multiple classes.
\newblock In {\em Artificial Intelligence and Statistics}, pages 1095--1103,
  2014.

\bibitem{Zhu2017Unpaired}
J.-Y. Zhu, T.~Park, P.~Isola, and A.~A. Efros.
\newblock Unpaired image-to-image translation using cycle-consistent
  adversarial networks.
\newblock {\em arXiv preprint arXiv:1703.10593}, 2017.

\bibitem{Zhuang2015Supervised}
F.~Zhuang, X.~Cheng, P.~Luo, S.~J. Pan, and Q.~He.
\newblock Supervised representation learning: Transfer learning with deep
  autoencoders.
\newblock In {\em IJCAI}, pages 4119--4125, 2015.

\end{thebibliography}
}

\end{document}